%% file: arxiv_main.tex
\documentclass[10pt,twocolumn,letterpaper]{article}

\usepackage{titling}
\usepackage[pagenumbers]{cvpr} 
\usepackage{graphicx}
\usepackage{amsmath}
\usepackage{amssymb}
\usepackage{booktabs}
\usepackage{multirow}
\usepackage{bbm}
\usepackage{soul}      
\usepackage{relsize}    
\usepackage{bbding}   
\usepackage[table]{xcolor}
\usepackage[pagebackref,breaklinks,colorlinks]{hyperref}
\usepackage[capitalize]{cleveref}

\crefname{section}{Sec.}{Secs.}
\Crefname{section}{Section}{Sections}
\Crefname{table}{Table}{Tables}
\crefname{table}{Tab.}{Tabs.}

\definecolor{darkblue}{rgb}{0.1, 0.3, 0.7}
\definecolor{orange}{rgb}{1.0, 0.5, 0.0}
\definecolor{red}{rgb}{1.0, 0.0, 0.0}
\definecolor{purple}{rgb}{0.6, 0.0, 0.6}

\def\cvprPaperID{0000}
\def\confName{CVPR}
\def\confYear{2022}

\begin{document}
\title{Collaborative Transformers for Grounded Situation Recognition}
\author{
Junhyeong Cho$^1$ \qquad Youngseok Yoon$^1$ \qquad Suha Kwak$^{1,2}$\\
Department of CSE, POSTECH$^1$ \qquad Graduate School of AI, POSTECH$^2$\\
{\tt\small \{junhyeong99, yys8646, suha.kwak\}@postech.ac.kr}
}

\maketitle

\input{arxiv_0_abstract}
\input{arxiv_1_introduction}
\input{arxiv_2_related_work}
\input{arxiv_3_method}
\input{arxiv_4_experiments}
\input{arxiv_5_conclusion}
\input{arxiv_6_acknowledgement}

{\small
\bibliographystyle{ieee_fullname}
\bibliography{arxiv_main}
}

\clearpage
\setcounter{section}{0}
\setcounter{figure}{0}
\setcounter{table}{0}
\renewcommand\thesection{\Alph{section}}
\renewcommand\thesubsection{\thesection.\arabic{subsection}}

\title{Collaborative Transformers for Grounded Situation Recognition\\
\vspace{2mm}
\textmd{--- Supplementary Material ---}
\vspace{-12mm}}
\author{}

\maketitle

\input{arxiv_supp_0_intro}
\input{arxiv_supp_1_method}
\input{arxiv_supp_2_implementation}
\input{arxiv_supp_3_qualitative}
\input{arxiv_supp_4_application}

\end{document}

%% file: arxiv_0_abstract.tex
\begin{abstract}
Grounded situation recognition is the task of predicting the main activity, entities playing certain roles within the activity, and bounding-box groundings of the entities in the given image.
To effectively deal with this challenging task, we introduce a novel approach where the two processes for activity classification and entity estimation are interactive and complementary.
To implement this idea, we propose \textbf{Co}llaborative Glance-Gaze Trans\textbf{Former} (CoFormer) that consists of two modules: Glance transformer for activity classification and Gaze transformer for entity estimation.
Glance transformer predicts the main activity with the help of Gaze transformer that analyzes entities and their relations, while Gaze transformer estimates the grounded entities by focusing only on the entities relevant to the activity predicted by Glance transformer.
Our CoFormer achieves the state of the art in all evaluation metrics on the SWiG dataset.
Training code and model weights are
available at \url{https://github.com/jhcho99/CoFormer}.
\end{abstract}

%% file: arxiv_1_introduction.tex
\section{Introduction}
\label{sec:intro}
\begin{figure}[t!]
    \centering
    \includegraphics[width=\columnwidth]{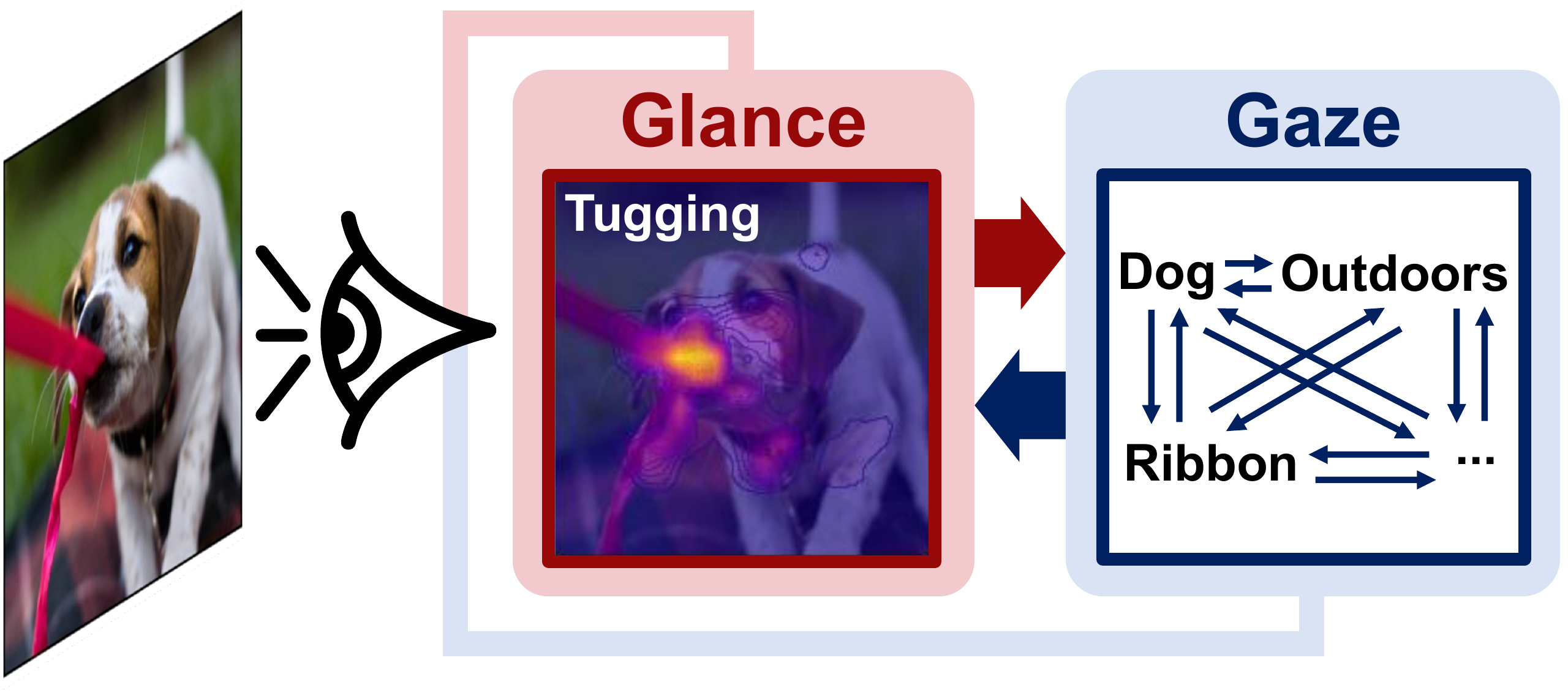}
    \caption{
    Two processes in comprehensive scene understanding.
    \textit{Glance} figures out what is happening, and \textit{Gaze} analyzes entities engaged in the main activity and their relations.
    In our CoFormer,
    these two processes are interactive and complementary.
    }
    \label{fig:overview}
\end{figure}

Humans make decisions via dual systems of thinking as stated in the cognitive theory by Kahneman~\cite{kahneman2003maps}.
Those two systems are known to work in tandem and complement each other~\cite{Peters06_numeracy,HabitFormation}.
Consider a comprehensive scene understanding task as a specific example of such decision making.
As illustrated in Figure~\ref{fig:overview},
humans cast a quick glance to figure out what is happening, and slowly gaze at details to analyze which objects are involved and how they are related.
These two processes are mutually supportive, \eg, understanding involved objects and their relations leads to more accurate recognition of the event depicted in the scene.

\begin{figure}[t!]
    \centering
    \includegraphics[width=\columnwidth]{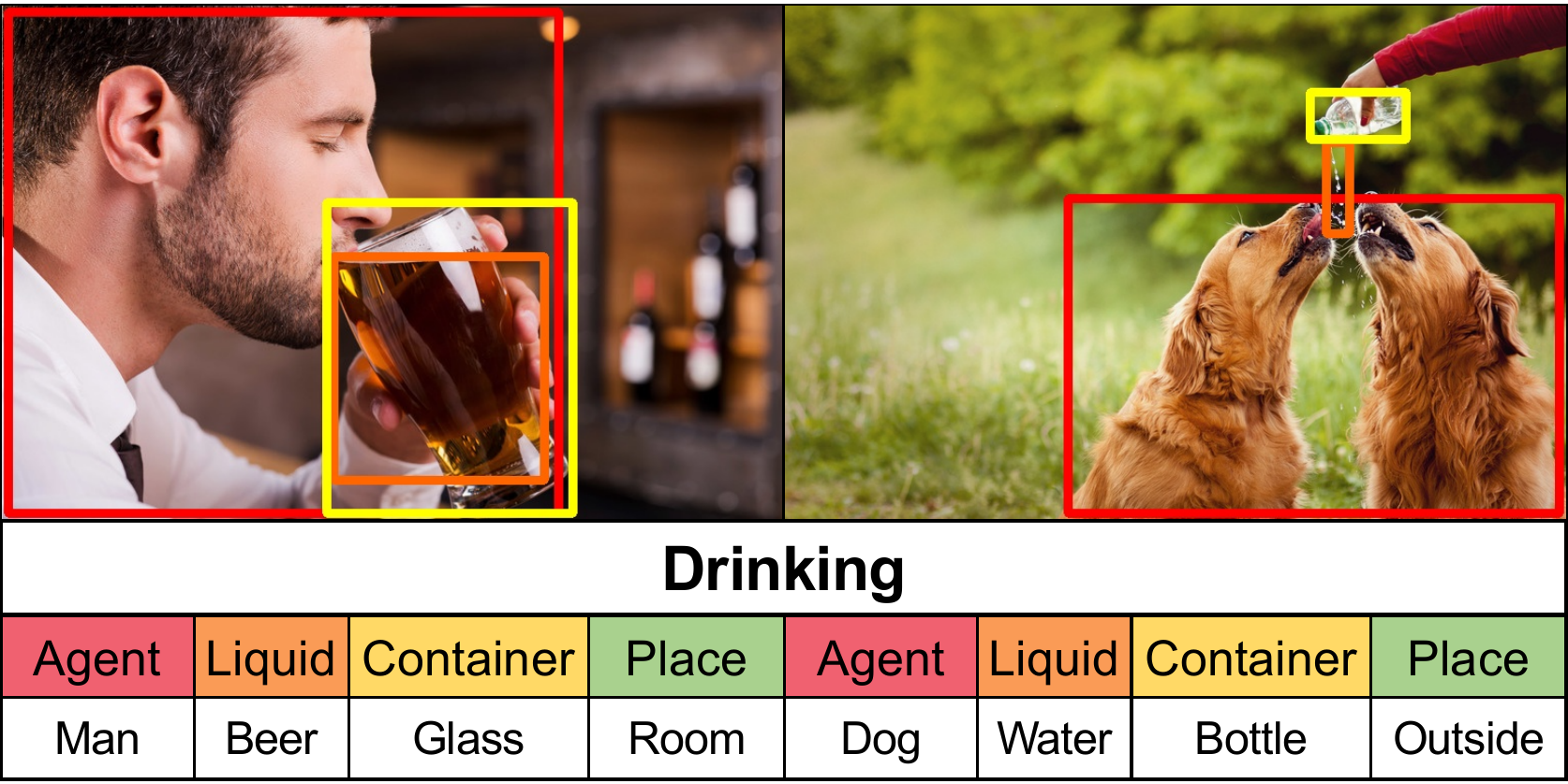}
    \caption{
    Two examples of Grounded Situation Recognition~\cite{pratt2020grounded}.
    These show various situations for the same verb.
    }
    \vspace{-1mm}
    \label{fig:dataset}
\end{figure}

Inspired by this, we propose a collaborative framework 
which leverages the two processes
for \mbox{Grounded} \mbox{Situation} \mbox{Recognition} (GSR)~\cite{pratt2020grounded}.
GSR is a comprehensive scene understanding task that is recently introduced as an extension of Situation Recognition (SR)~\cite{yatskar2016situation}.
The objective of SR is to produce a structured image summary that describes the main activity and entities 
playing certain roles within the activity,
where the roles are predefined for each activity by a lexical database called FrameNet~\cite{fillmore2003background}.
In GSR, those involved entities are grounded with bounding boxes;
Figure~\ref{fig:dataset} presents example results of GSR.
Following conventions, we call an activity \emph{verb} and an entity \emph{noun} in this paper.

The common pipeline of SR and GSR in the
literature~\cite{yatskar2016situation,yatskar2017commonly,mallya2017recurrent,li2017situation,suhail2019mixture,cooray2020attention,pratt2020grounded,cho2021gsrtr} resembles the two processes:
predicting a verb (\emph{Glance}), then estimating a noun for each role 
associated with the predicted verb \mbox{(\emph{Gaze}).
Regarding} this pipeline, correctness of the predicted verb is extremely important since noun estimation entirely depends on the predicted verb.
If the result of verb prediction is incorrect, then estimated nouns
cannot be correct either because the predicted verb determines the set of roles, \ie, the basis of noun estimation.
Moreover, verb prediction is challenging since a verb is highly abstract 
and situations for the same verb could significantly vary as shown in \mbox{Figure}~\ref{fig:dataset}.
In spite of
its importance and difficulty, verb prediction has been made in na\"ive ways, \eg, using a single classifier on top of a convolutional neural network (CNN), which is analogous to \emph{Glance} only.
Existing methods allow \emph{Glance} to assist \emph{Gaze} by informing the predicted verb but not vice versa; this could limit the performance of verb prediction, and consequently, that of the entire pipeline.

We resolve the above issue by a collaborative framework that enables \emph{Glance} and \emph{Gaze} to interact and complement each other. 
To fully utilize this framework,
we propose 
\textbf{Co}llaborative Glance-Gaze Trans\textbf{Former} (CoFormer)
that consists of \mbox{Glance} transformer and \mbox{Gaze} transformer
as illustrated in Figure~\ref{fig:overall}.
\mbox{Glance transformer} predicts a verb by aggregating image features 
through self-attentions, and Gaze transformer estimates nouns and their groundings by allowing each role to focus on its relevant image region through 
\mbox{self-attentions} and \mbox{cross-attentions}.
As shown in Figure~\ref{fig:overall}, 
there are two steps for \emph{Gaze} in our \mbox{CoFormer}.
\mbox{Gaze-Step1} transformer estimates nouns for all role candidates and assists \mbox{Glance} transformer for more accurate verb prediction.
Meanwhile,
\mbox{Gaze-Step2} transformer estimates a noun and its grounding for each role associated with the predicted verb by exploiting the aggregated image features obtained by \mbox{Glance} transformer.

The collaborative relationship between \mbox{Glance} and \mbox{Gaze} transformers lead to more accurate verb and grounded noun predictions for GSR.
In CoFormer,
\mbox{\emph{Gaze-Step1}} supports \emph{Glance} by analyzing
involved nouns and their relations, 
which enables \mbox{noun-aware} verb prediction.
\emph{Glance} assists \mbox{\emph{Gaze-Step2}} by informing the predicted verb, which reduces the role candidates considered in grounded noun prediction.

\noindent \textbf{Contributions.} 
\textbf{(i)} We propose a collaborative framework where the two processes for verb prediction and noun estimation are interactive and complementary, which is novel in GSR.
\textbf{(ii)} Our method achieves \mbox{state-of-the-art} accuracy
in every evaluation metric on the SWiG dataset.
\textbf{(iii)} We demonstrate the effectiveness of \mbox{CoFormer}
by conducting extensive experiments and provide \mbox{in-depth} analyses.

%% file: arxiv_2_related_work.tex
\begin{figure*}[t!]
    \centering
    \includegraphics[width=\textwidth]{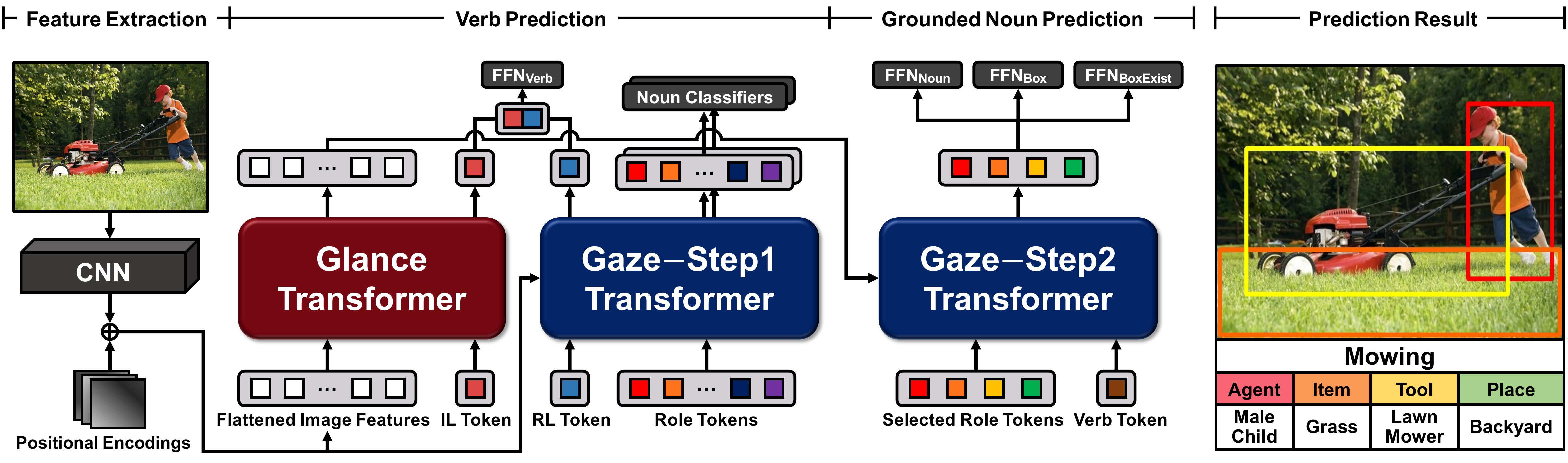}
    \caption{
    Overall architecture of \textbf{Co}llaborative Glance-Gaze Trans\textbf{Former} (CoFormer). 
    Glance transformer predicts a verb with the help of \mbox{Gaze-Step1} transformer that analyzes nouns and their relations by leveraging role features, while \mbox{Gaze-Step2} transformer estimates the grounded nouns for the roles associated with the predicted verb.
    Prediction results are obtained by feed forward networks (FFNs).
    The results from the two noun classifiers placed on top of \mbox{Gaze-Step1} transformer are ignored at inference time.
    }
    \label{fig:overall}
\end{figure*}

\section{Related Work}
\label{sec:related}
Visual reasoning such as 
image captioning~\cite{vinyals2015show,you2016image,huang2019attention,guo2020normalized,chen2021human}, scene graph generation~\cite{xu2017scene,yang2018graph,khandelwal2021segmentation,lu2021context}, and \mbox{human-object-interaction} detection~\cite{li2019transferable,wang2020learning,kim2021hotr,zhang2021spatially} 
has been widely studied for comprehensive understanding of images.
Given an image, image captioning aims at describing activities and entities using natural language, and scene graph generation or \mbox{human-object-interaction} detection aims at capturing a set of triplets
\mbox{$\langle \mathrm{subject, \; predicate, \; object} \rangle$} or \mbox{$\langle \mathrm{human, \; object, \; interaction} \rangle$}.
However, it is not straightforward to evaluate the quality of natural language captions, and the triplets have limited expressive power.
To overcome such limitations, Yatskar \etal~\cite{yatskar2016situation} introduce SR along with the \textit{imSitu} dataset.
SR has more expressive power based on linguistic sources from FrameNet~\cite{fillmore2003background}, and its quality evaluation is straightforward.
GSR builds upon SR by additionally estimating \mbox{bounding-box} groundings.

\noindent \textbf{Situation Recognition.} 
Yatskar \etal~\cite{yatskar2016situation} propose a conditional random field~\cite{lafferty2001conditional} model, and also present a tensor composition method with semantic augmentation~\cite{yatskar2017commonly}. 
Mallya and Lazebnik~\cite{mallya2017recurrent} employ a recurrent neural network to capture role relations in the predefined sequential order. 
Li \etal~\cite{li2017situation} propose a gated graph neural network (GGNN)~\cite{li2016gated} to capture the relations in more flexible ways.
To learn \mbox{context-aware} role relations depending on an input image, Suhail and Sigal~\cite{suhail2019mixture} apply a mixture kernel method to GGNN.
Cooray \etal~\cite{cooray2020attention} employ \mbox{inter-dependent} queries to capture role relations, and present a verb model which considers
nouns from the two predefined roles;
they construct a query based on two nouns for verb prediction.
Compared with this, CoFormer considers nouns from all role candidates for accurate verb prediction. 

\noindent \textbf{Grounded Situation Recognition.} 
Pratt \etal~\cite{pratt2020grounded} propose GSR along with the \emph{SWiG} dataset, and present two models: Independent Situation Localizer (ISL) and Joint Situation Localizer (JSL).
They first predict a verb using a single classifier on top of a CNN backbone,
then estimate nouns and their groundings.
In both models, LSTM~\cite{hochreiter1997long} produces output features to predict nouns in the predefined sequential order, while RetinaNet~\cite{lin2017focal} estimates their groundings.
ISL separately predicts nouns and their groundings, and JSL jointly predicts them. 
Cho \etal~\cite{cho2021gsrtr} propose a transformer \mbox{encoder-decoder} architecture,
where the encoder effectively captures \mbox{high-level} semantic features for verb prediction and the decoder flexibly learns the role \mbox{relations.
Compared} with these models,
CoFormer leverages involved nouns and their relations for accurate verb prediction via transformers.

\noindent \textbf{Transformer Architecture.} 
Transformers~\cite{vaswani2017attention} have driven remarkable success in vision tasks~\cite{carion2020end, guo2020normalized, dosovitskiy2021an, chen2021human, lu2021context, kim2021hotr, liu2021paint, lee2021ctrl}.
Dosovitskiy \etal~\cite{dosovitskiy2021an} propose a transformer encoder architecture for image classification by aggregating image features using a learnable token in the encoder.
Carion \etal~\cite{carion2020end} present a transformer encoder-decoder architecture for object detection by predicting a set of bounding boxes using a fixed number of learnable queries in the decoder.
Such learnable queries have been widely used to extract features in other transformer architectures~\cite{kim2021hotr, liu2021paint, lee2021ctrl}.
Compared with those transformers, CoFormer employs two learnable tokens which aggregate different kinds of features through self-attentions. 
In addition, CoFormer constructs a different number of learnable queries by explicitly leveraging the prediction result obtained by two encoders and a classifier. 

%% file: arxiv_3_method.tex
\section{Method}
\label{sec:method}
\noindent \textbf{Task Definition.} 
GSR assumes discrete sets of verbs $\mathcal V$, nouns $\mathcal N$, and roles $\mathcal R$. 
Each verb $v \in \mathcal V$ is paired with a frame derived from FrameNet~\cite{fillmore2003background}, where the frame defines the set of roles $\mathcal R_v \subset \mathcal R$ associated with the verb.
For example, a verb \textit{Mowing} is paired with a frame which defines the set of roles $\mathcal R_{\mathrm{\textit{Mowing}}} = \{\mathrm{\textit{Agent}}, \mathrm{\textit{Item}}, \mathrm{\textit{Tool}}, \mathrm{\textit{Place}}\}$ as shown in Figure~\ref{fig:overall}.
Each role $r \in \mathcal R_v$ is fulfilled by a noun $n \in \mathcal N$ grounded by a bounding box $\mathbf b \in \mathbb R^4$, called \textit{grounded noun}.
Formally speaking, the set of fulfilled roles is $\mathcal F_v = \{(r_i, n_i, \mathbf b_i) \; \vert \; r_i \in \mathcal R_v, \; n_i \in \mathcal N \cup \{ \emptyset_n \}, \; \mathbf b_i \in \mathbb R^4 \cup \{ \emptyset_b \}$ for $i=1,...,\vert \mathcal R_v \vert \}$; $\emptyset_n$ and $\emptyset_b$ denote \textit{unknown} and \textit{not grounded}, respectively.
The output of GSR is a grounded situation denoted by $S = (v, \mathcal F_v)$.

\subsection{Overall Architecture}
\begin{figure*}[t!]
    \centering
    \includegraphics[width=\textwidth]{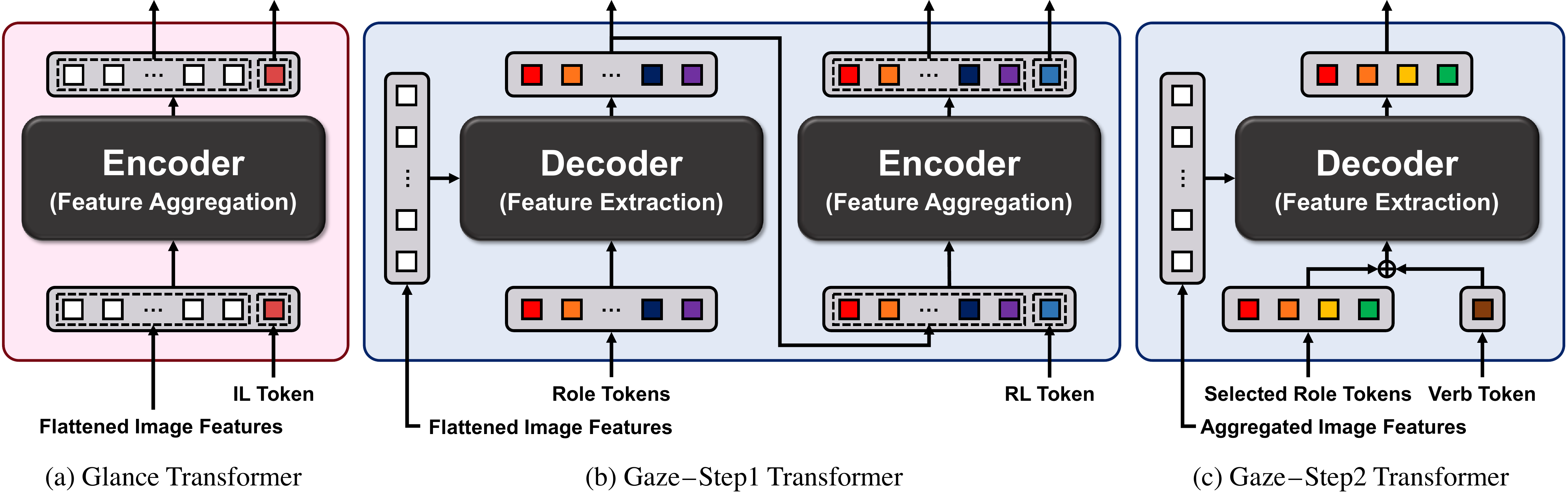}
    \caption{
    Transformer architectures in \mbox{CoFormer} are composed of
    common building blocks, encoder and decoder.
    }
    \label{fig:transformer}
\end{figure*}

\mbox{CoFormer} predicts a verb, then estimates grounded nouns as illustrated in Figure~\ref{fig:overall}.
As shown in Figure~\ref{fig:transformer}, our transformers consist of common building blocks, encoder and decoder, whose architectures are illustrated
in Figure~\ref{fig:block}.
For simplicity, we abbreviate Step1 as \emph{S1}, and Step2 as \emph{S2} in the remaining of this paper. 

\begin{figure}[!t]
    \centering
    \includegraphics[width=\columnwidth]{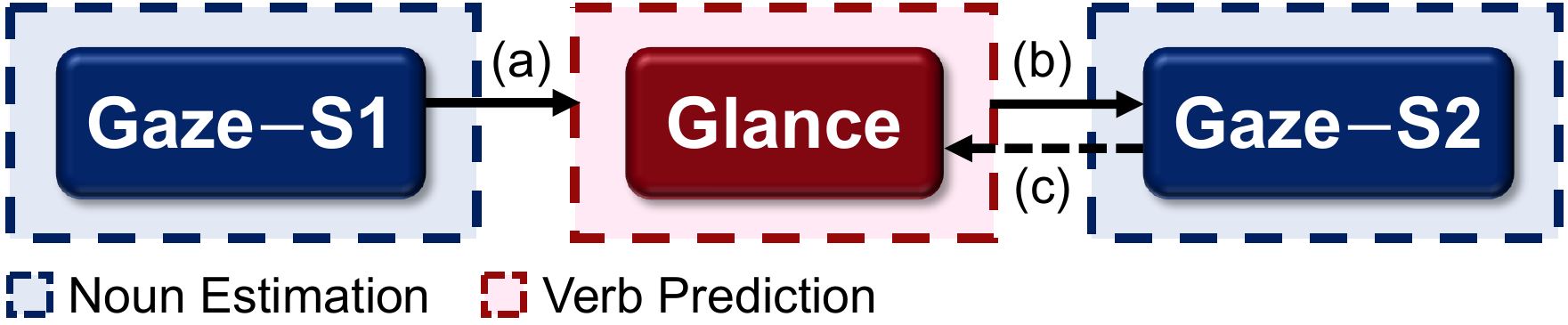}
    \vspace{-6.5mm}
    \caption{
    Interactive and complementary processes in \mbox{CoFormer}.
    (a) RL token feature, (b) predicted verb, (c) loss gradients. 
    }
    \vspace{-2.7mm}
    \label{fig:model_diagram}
\end{figure}

\noindent \textbf{Overview.}
Given an image, \mbox{CoFormer} extracts 
flattened
image features via a CNN backbone and flatten operation, which are fed as input to Glance transformer and \mbox{Gaze-S1} transformer. 
From these transformers, 
the output features corresponding to \mbox{Image-Looking} (IL) and \mbox{Role-Looking} (RL) tokens are used for verb prediction.
Considering the predicted verb, \mbox{Gaze-S2} transformer estimates grounded nouns for the roles associated with the predicted verb 
by exploiting image features obtained by
Glance transformer.
Figure~\ref{fig:model_diagram} shows the collaborative relationship between the modules;
transformers for verb prediction and noun estimation 
are interactive and complementary 
in \mbox{CoFormer}.

\noindent \textbf{Glance Transformer.}
This transformer consists of a single encoder which takes the flattened image features and learnable IL token as input.
IL token captures the essential features for verb prediction, while Glance transformer aggregates the image features through self-attentions.

\noindent \textbf{Gaze-S1 Transformer.}
This transformer is composed of a decoder and an encoder.
The decoder takes the flattened image features and learnable role tokens as input, where the role tokens correspond to all role candidates.
This module extracts role features from the image features via the role tokens.
Then, the encoder takes the role features and learnable RL token as input.
RL token captures involved nouns and their relations for verb prediction, while the encoder aggregates the role features through \mbox{self-attentions}.

\noindent \textbf{Gaze-S2 Transformer.}
This transformer consists of a single decoder, which takes learnable tokens and
the aggregated image features obtained from Glance transformer as input.
The input tokens correspond to the predicted verb and its associated roles.
Note that a verb token is added to role tokens as shown in Figure~\ref{fig:transformer};
conditioning on the predicted verb 
significantly reduces the search space 
of the roles,
\eg, the search space of \textit{Mowing Tool} is much smaller than that of \textit{Tool}. 
\mbox{Gaze-S2} transformer \mbox{extracts role} features from the aggregated image features,
and
the extracted role features are used for grounded noun prediction.

\subsection{Feature Extraction}
\begin{figure}[t!]
    \centering
    \includegraphics[width=\columnwidth]{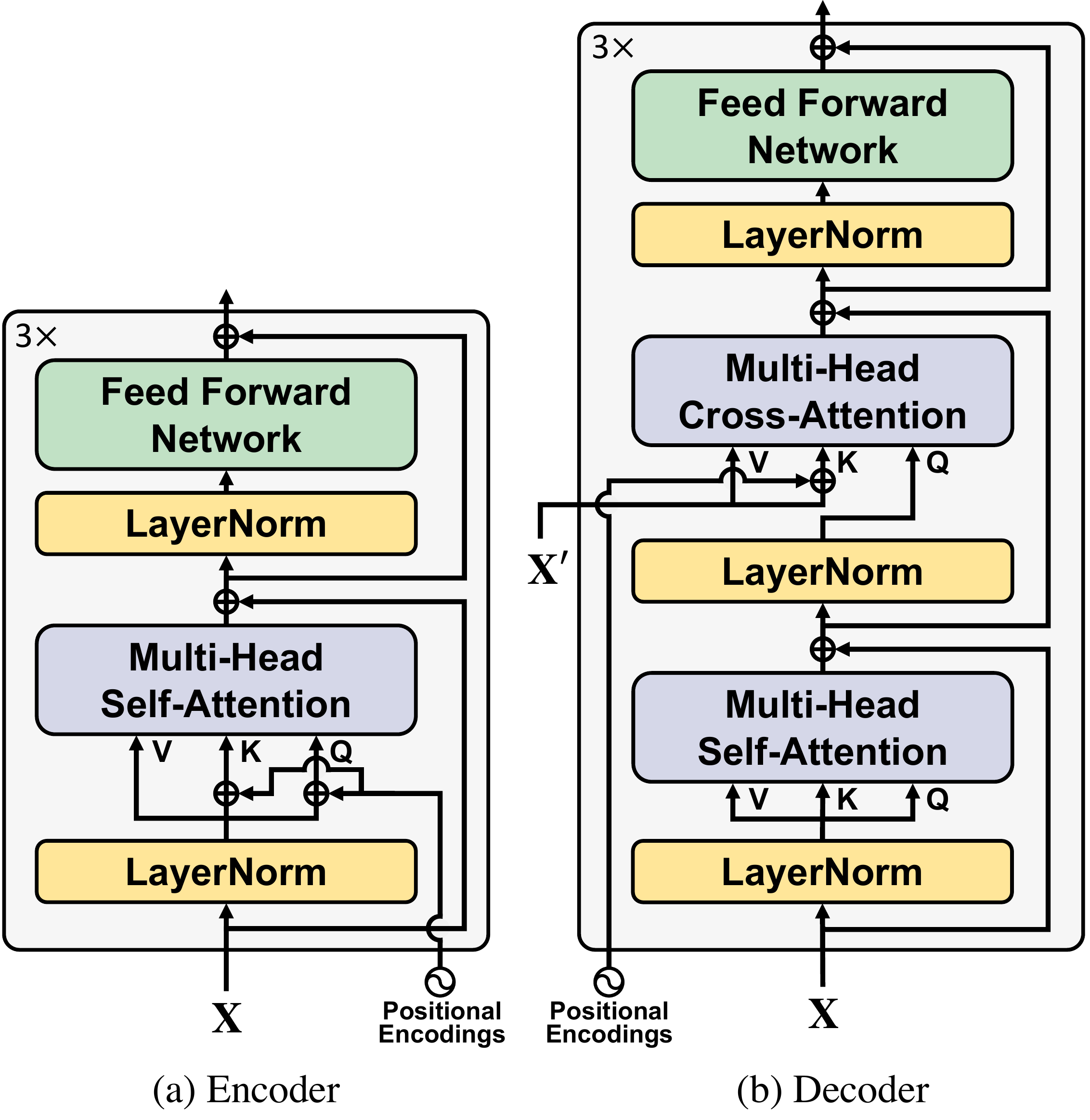}
    \caption{
    Detailed architectures of encoder and decoder.
    We use \mbox{Pre-Layer} Normalization~\cite{xiong2020layer} for these two modules.
    An encoder performs feature aggregation through \mbox{self-attentions} on $\mathbf X$, and a decoder performs feature extraction through \mbox{self-attentions} on $\mathbf X$ and \mbox{cross-attentions} between $\mathbf X$ and $\mathbf X'$.
    }
    \label{fig:block}
\end{figure}

Given an input image, a single CNN backbone extracts image features of size $h \times w \times c$, where $h \times w$ is the resolution, and $c$ is the number of channels.
Then, a $1 \times 1$ convolution followed by a flatten operation produces flattened image features $\mathbf X_{F} \in \mathbb R^{hw \times d}$, where $d$ is the number of channels. 
The flattened image features $\mathbf X_{F}$ are fed as input to Glance transformer (Fig.~\ref{fig:transformer}(a)) and \mbox{Gaze-S1} transformer (Fig.~\ref{fig:transformer}(b)).
For the flattened image features $\mathbf X_{F}$, positional encodings are introduced to retain spatial information.
As shown in Figure~\ref{fig:block}, positional encodings are added to the queries and keys at the \mbox{self-attention} layers in an encoder, and to the keys at the \mbox{cross-attention} layers in a decoder.

\subsection{Verb Prediction}
The input of the encoder in Glance transformer is obtained by the concatenation of the image features $\mathbf X_{F}$ and learnable IL token.
IL token captures the essential features for verb prediction, while the encoder aggregates the image features through \mbox{self-attentions}.
As its output, the encoder produces aggregated image features $\mathbf X_{A} \in \mathbb R^{hw \times d}$ and IL token feature.
For the aggregated image features $\mathbf X_{A}$, positional encodings are 
applied.

\mbox{Gaze-S1} transformer supports Glance transformer for more accurate verb prediction, while predicting nouns for all role candidates.
To be specific, the decoder of \mbox{Gaze-S1} transformer takes the flattened image features $\mathbf X_{F}$ and learnable role tokens corresponding to all predefined roles; each role token embedding is denoted by $\mathbf w_r \in \mathbb R^{d}$, where $r \in \mathcal R$.
This decoder extracts role features through self-attentions on the role tokens and cross-attentions between the tokens and the image features $\mathbf X_{F}$.
The input of the encoder in \mbox{Gaze-S1} transformer is obtained by the concatenation of the extracted role features and learnable RL token.
RL token captures involved nouns
and their relations from all role candidates, while the encoder aggregates the role features through self-attentions.
For this encoder, positional encodings are not added to the queries and keys at the self-attention layers since roles are \mbox{permutation-invariant} in GSR.
Regarding to \mbox{Gaze-S1} transformer, the extracted and aggregated role features are fed as input to noun classifiers; these classifiers are auxiliary modules and their results
are ignored at inference time.
Note that \mbox{Gaze-S1} transformer assists Glance transformer via
RL token feature which is aware of involved nouns and their relations.

IL token feature and RL token feature are concatenated, then fed as input to the feed forward network (FFN) for verb classification, which consists of learnable linear layers with activation functions.
The verb classifier $\mathrm{FFN}_{\mathrm {Verb}}$ followed by a softmax function produces a verb probability distribution $\mathbf p_v$, which is used to estimate the most probable verb $\hat v= \arg \max_v \mathbf p_v$.
The predicted verb $\hat v$ supports \mbox{Gaze-S2} transformer so that the transformer concentrates only on the roles associated with the predicted verb and estimates their grounded nouns more accurately in consequence.

\subsection{Grounded Noun Prediction}
The aggregated image features $\mathbf X_{A}$ from Glance transformer are fed as input to \mbox{Gaze-S2} transformer (Fig.~\ref{fig:transformer}(c)).
The decoder in this transformer takes the image features $\mathbf X_{A}$ and \mbox{\textit{frame-role queries}} as input.
Specifically, for each role $r$ in the frame of the predicted verb $\hat v$, its \mbox{frame-role} query $\mathbf q_r \in \mathbb R^{d}$ is constructed by an addition of the learnable role token embedding $\mathbf w_r \in \mathbb R^{d}$ and the learnable verb token embedding $\mathbf w_{\hat v} \in \mathbb R^{d}$, \ie, $\mathbf q_r = \mathbf w_r + \mathbf w_{\hat v}$ for $r \in \mathcal R_{\hat v}$.
The decoder extracts role features through \mbox{self-attentions} on the \mbox{frame-role} queries and \mbox{cross-attentions} between the queries and the image features $\mathbf X_{A}$ to capture the involved nouns and their relations from roles relevant to the verb $\hat v$.
Those extracted role features are used for grounded noun prediction.
Note that this task requires to predict a noun, a bounding box, and a box existence for each role.
Accordingly,
we employ three feed forward networks $\mathrm{FFN}_{\mathrm {Noun}}$, $\mathrm{FFN}_{\mathrm {Box}}$, and $\mathrm{FFN}_{\mathrm {BoxExist}}$ that take the role features as input for noun classification, bounding box estimation, and box existence prediction, respectively.
Each of these FFNs consists of learnable linear layers with activation functions.

For each role $r \in \mathcal R_{\hat v}$, $\mathrm{FFN}_{\mathrm {Noun}}$ followed by a softmax function produces a noun probability distribution $\mathbf p_{n_r}$.
$\mathrm{FFN}_{\mathrm {Box}}$ followed by a sigmoid function produces a bounding box $\hat{\mathbf b}_r \in [0,1]^4$ which indicates the center coordinates, height and width relative to the input image size.
The predicted box $\hat{\mathbf b}_r$ can be
transformed into the \mbox{top-left} and \mbox{bottom-right} coordinates  $\hat{\mathbf b}'_r \in \mathbb R^4$. 
$\mathrm{FFN}_{\mathrm {Box Exist}}$ followed by a sigmoid function produces a box existence probability $p_{b_r} \in [0, 1]$. 
If $p_{b_r} < 0.5$, 
the predicted box $\hat{\mathbf b}'_r$ 
is ignored.
Note that the predicted verb $\hat v$ assists \mbox{Gaze-S2} transformer via the construction of \mbox{frame-role} queries, while the loss gradients propagated from \mbox{Gaze-S2} transformer through the aggregated image features $\mathbf X_{A}$ enable Glance transformer to implicitly consider involved nouns.

\subsection{Training \mbox{CoFormer}}
The predicted verb, nouns and bounding boxes are used for computing losses to train \mbox{CoFormer}. 
At training time, we construct \mbox{frame-role} queries based on the \mbox{ground-truth} verb for stable training of \mbox{Gaze-S2} transformer.
Please refer to the supplementary material for more training details.

\noindent \textbf{Verb Classification Loss.}
The verb classification loss is the \mbox{cross-entropy} between the verb probability distribution $\mathbf p_v$ and the \mbox{ground-truth} verb distribution. 

\noindent \textbf{Noun Classification Losses.}
As illustrated in Figure~\ref{fig:overall}, \mbox{CoFormer} has three noun classifiers; two of them are placed on top of \mbox{Gaze-S1} transformer and the other is incorporated with \mbox{Gaze-S2} transformer. 
For each noun classifier, we compute the \mbox{cross-entropy} between the estimated noun probability distribution and the \mbox{ground-truth} noun distribution for each role $r \in \mathcal R_{\tilde v}$, where $\tilde v$ is the \mbox{ground-truth} verb.
The computed \mbox{cross-entropy} loss is averaged over roles $R_{\tilde v}$.
Note that we only train role tokens for the roles in the frame of the \mbox{ground-truth} verb $\tilde v$, since noun annotations are given for the roles associated with the verb $\tilde v$ in the dataset.

\noindent \textbf{Box Existence Prediction Loss.}
To deal with roles 
which have no \mbox{ground-truth} boxes (\ie, $\emptyset_b$), \eg, by occlusion, \mbox{CoFormer} estimates a box existence probability $p_{b_r}$ for each role $r \in \mathcal R_{\tilde v}$.
The box existence prediction loss is the cross-entropy between the probability $p_{b_r}$ and the \mbox{ground-truth} box existence, which is averaged over roles $R_{\tilde v}$.

\noindent \textbf{Box Regression Losses.}
We employ $L1$ loss and $\mathrm {GIoU}$ loss~\cite{rezatofighi2019generalized} for box regression.
Let $\mathbf b_r$ denote the \mbox{ground-truth} box in the form of the center coordinates, height and width relative to the given image size.
In the computation of box regression losses, we ignore the roles which have no
\mbox{ground-truth} boxes (\ie, $\emptyset_b$).
The $L1$ box regression loss $\mathcal L_{L1}$ is computed by
\begin{align}
    \label{eq:loss_l1}
    \mathcal L_{L1} &= \frac{1}{\vert \tilde{\mathcal R} \vert} \sum_{r \in \tilde{\mathcal R}} \Vert \mathbf b_r - \hat{\mathbf b}_r \Vert_1,
\end{align}
where $\tilde{\mathcal R} = \{ r \; \vert \; \mathbf b_r \neq \emptyset_b$ for $r \in R_{\tilde v} \}$.
To compute the $\mathrm {GIoU}$ loss, 
$\mathrm{GIoU}(\cdot$)
is first computed by
\begin{align}
    \label{eq:giou}
    \mathrm{GIo}&\mathrm{U}(\mathbf b'_r, \hat{\mathbf b}'_r) \nonumber \\
        &= \frac{\vert \mathbf b'_r \cap \hat{\mathbf b}'_r \vert}{\vert \mathbf b'_r \cup \hat{\mathbf b}'_r \vert} - \frac{\vert C(\mathbf b'_r, \hat{\mathbf b}'_r) \setminus (\mathbf b'_r \cup \hat{\mathbf b}'_r) \vert}{\vert C(\mathbf b'_r, \hat{\mathbf b}'_r)\vert},
\end{align}
where $\mathbf b'_r$ indicates the \mbox{top-left} and \mbox{bottom-right} coordinates transformed from $\mathbf b_r$, and $C(\mathbf b'_r, \hat{\mathbf b}'_r)$ denotes the smallest box which encloses $\mathbf b'_r$ and $\hat{\mathbf b}'_r$. The $\mathrm {GIoU}$ box regression loss $\mathcal L_{\mathrm {GIoU}}$ is then computed by
\begin{align}
    \label{eq:loss_giou}
    \mathcal L_{\mathrm{GIoU}} =
    &\frac{1}{\vert \tilde{\mathcal R} \vert} \sum_{r\in \tilde{\mathcal R}}
    \left( 1 - \mathrm{GIoU}(\mathbf b'_r, \hat{\mathbf b}'_r) \right).
\end{align}

%% file: arxiv_4_experiments.tex
\section{Experiments}
\label{sec:exp}

\begin{table*}[!t]
    \centering
    \resizebox{\textwidth}{!}{
        \begin{tabular}{l|l|ccccc|ccccc|cccc}
        \hline
        \multicolumn{2}{c|}{}
            & \multicolumn{5}{c|}{Top-1 Predicted Verb}
            & \multicolumn{5}{c|}{Top-5 Predicted Verbs}
            & \multicolumn{4}{c}{Ground-Truth Verb}  
        \\
        \hline
            &  
            &       &       &       & grnd & grnd
            &       &       &       & grnd & grnd
            &       &       & grnd & grnd  
        \\
        Set & Method
            & verb & value & value-all & value & value-all
            & verb & value & value-all & value & value-all
            & value & value-all & value & value-all
        \\
        \hline
        \hline
            & \multicolumn{15}{c}{\textit{Methods for Situation Recognition}}
        \\
        \cline{2-16}
        \multirow{9.5}{*}{dev}
            & CRF \cite{yatskar2016situation}
            & 32.25 & 24.56 & 14.28 & -- & -- 
            & 58.64 & 42.68 & 22.75 & -- & -- 
            & 65.90 & 29.50 & -- & -- 
        \\
            & CRF w/ DataAug \cite{yatskar2017commonly}
            & 34.20 & 26.56 & 15.61 & -- & --
            & 62.21 & 46.72 & 25.66 & -- & --
            & 70.80 & 34.82 & -- & --
        \\
            & RNN w/ Fusion \cite{mallya2017recurrent}
            & 36.11 & 27.74 & 16.60 & -- & --
            & 63.11 & 47.09 & 26.48 & -- & --
            & 70.48 & 35.56 & -- & --
        \\
            & GraphNet \cite{li2017situation}
            & 36.93 & 27.52 & 19.15 & -- & --
            & 61.80 & 45.23 & 29.98 & -- & --
            & 68.89 & 41.07 & -- & --
        \\
            & CAQ w/ RE-VGG \cite{cooray2020attention}
            & 37.96 & 30.15 & 18.58 & -- & --
            & 64.99 & 50.30 & 29.17 & -- & --
            & \textbf{73.62} & 38.71 & -- & --
        \\
            & Kernel GraphNet \cite{suhail2019mixture}
            & \textbf{43.21} & \textbf{35.18} & \textbf{19.46} & -- & --
            & \textbf{68.55} & \textbf{56.32} & \textbf{30.56} & -- & --
            & 73.14 & \textbf{41.68} & -- & --
        \\
        \cline{2-16}
            & \multicolumn{15}{c}{\textit{Methods for Grounded Situation Recognition}}
        \\
        \cline{2-16}
            & ISL \cite{pratt2020grounded}
            & 38.83 & 30.47 & 18.23 & 22.47 & 7.64
            & 65.74 & 50.29 & 28.59 & 36.90 & 11.66
            & 72.77 & 37.49 & 52.92 & 15.00 
        \\
            & JSL \cite{pratt2020grounded}
            & 39.60 & 31.18 & 18.85 & 25.03 & 10.16
            & 67.71 & 52.06 & 29.73 & 41.25 & 15.07
            & 73.53 & 38.32 & 57.50 & 19.29
        \\
            & GSRTR \cite{cho2021gsrtr}
            & 41.06 & 32.52 & 19.63 & 26.04 & 10.44
            & 69.46 & 53.69 & 30.66 & 42.61 & 15.98
            & 74.27 & 39.24 & 58.33 & 20.19
        \\
            & \cellcolor[gray]{0.9}\mbox{CoFormer} (Ours)
            & \cellcolor[gray]{0.9}\textbf{44.41} & \cellcolor[gray]{0.9}\textbf{35.87} & \cellcolor[gray]{0.9}\textbf{22.47} & \cellcolor[gray]{0.9}\textbf{29.37} & \cellcolor[gray]{0.9}\textbf{12.94}
            & \cellcolor[gray]{0.9}\textbf{72.98} & \cellcolor[gray]{0.9}\textbf{57.58} & \cellcolor[gray]{0.9}\textbf{34.09} & \cellcolor[gray]{0.9}\textbf{46.70} & \cellcolor[gray]{0.9}\textbf{19.06} 
            & \cellcolor[gray]{0.9}\textbf{76.17} & \cellcolor[gray]{0.9}\textbf{42.11} & \cellcolor[gray]{0.9}\textbf{61.15} & \cellcolor[gray]{0.9}\textbf{23.09} 
        \\
        \hline
            & \multicolumn{15}{c}{\textit{Methods for Situation Recognition}}
        \\
        \cline{2-16}
        \multirow{9.5}{*}{test}
             & CRF \cite{yatskar2016situation}
             & 32.34 & 24.64 & 14.19 & -- & --
             & 58.88 & 42.76 & 22.55 & -- & --
             & 65.66 & 28.96 & -- & --
        \\
            & CRF w/ DataAug \cite{yatskar2017commonly}
            & 34.12 & 26.45 & 15.51 & -- & --
            & 62.59 & 46.88 & 25.46 & -- & --
            & 70.44 & 34.38 & -- & --
        \\
            & RNN w/ Fusion \cite{mallya2017recurrent}
            & 35.90 & 27.45 & 16.36 & -- & --
            & 63.08 & 46.88 & 26.06 & -- & --
            & 70.27 & 35.25 & -- & --
        \\
            & GraphNet \cite{li2017situation}
            & 36.72 & 27.52 & 19.25 & -- & --
            & 61.90 & 45.39 & 29.96 & -- & --
            & 69.16 & 41.36 & -- & --
        \\
            & CAQ w/ RE-VGG \cite{cooray2020attention}
            & 38.19 & 30.23 & 18.47 & -- & --
            & 65.05 & 50.21 & 28.93 & -- & --
            & \textbf{73.41} & 38.52 & -- & --
        \\
            & Kernel GraphNet \cite{suhail2019mixture}
            & \textbf{43.27} & \textbf{35.41} & \textbf{19.38} & -- & --
            & \textbf{68.72} & \textbf{55.62} & \textbf{30.29} & -- & --
            & 72.92 & \textbf{42.35} & -- & --
        \\
        \cline{2-16}
            & \multicolumn{15}{c}{\textit{Methods for Grounded Situation Recognition}}
        \\
        \cline{2-16}
            & ISL \cite{pratt2020grounded}
            & 39.36 & 30.09 & 18.62 & 22.73 & 7.72
            & 65.51 & 50.16 & 28.47 & 36.60 & 11.56
            & 72.42 & 37.10 & 52.19 & 14.58
        \\
            & JSL \cite{pratt2020grounded}
            & 39.94 & 31.44 & 18.87 & 24.86 & 9.66
            & 67.60 & 51.88 & 29.39 & 40.60 & 14.72
            & 73.21 & 37.82 & 56.57 & 18.45
        \\
            & GSRTR \cite{cho2021gsrtr}
            & 40.63 & 32.15 & 19.28 & 25.49 & 10.10
            & 69.81 & 54.13 & 31.01 & 42.50 & 15.88
            & 74.11 & 39.00 & 57.45 & 19.67
        \\
            & \cellcolor[gray]{0.9}\mbox{CoFormer} (Ours)
            & \cellcolor[gray]{0.9}\textbf{44.66} & \cellcolor[gray]{0.9}\textbf{35.98} & \cellcolor[gray]{0.9}\textbf{22.22} & \cellcolor[gray]{0.9}\textbf{29.05} & \cellcolor[gray]{0.9}\textbf{12.21} 
            & \cellcolor[gray]{0.9}\textbf{73.31} & \cellcolor[gray]{0.9}\textbf{57.76} & \cellcolor[gray]{0.9}\textbf{33.98} & \cellcolor[gray]{0.9}\textbf{46.25} & \cellcolor[gray]{0.9}\textbf{18.37} 
            & \cellcolor[gray]{0.9}\textbf{75.95} & \cellcolor[gray]{0.9}\textbf{41.87} & \cellcolor[gray]{0.9}\textbf{60.11} & \cellcolor[gray]{0.9}\textbf{22.12} 
        \\
        \hline
    \end{tabular}}
    \vspace{-1mm}
    \caption{
        Quantitative evaluations of methods in SR and GSR.
        SR models are evaluated on the imSitu dataset, and GSR models are evaluated on the SWiG dataset.
        The only difference between the two datasets is the existence of bounding box annotation.
    }
    \vspace{-1mm}
    \label{table:result}
\end{table*}

\begin{table*}[!t]
    \centering
    \resizebox{\textwidth}{!}{
        \begin{tabular}{l|ccc|ccc|cccc}
        \hline
        \multicolumn{1}{c|}{}
            & \multicolumn{3}{c|}{\;Top-1 Predicted Verb\;}
            & \multicolumn{3}{c|}{\;Top-5 Predicted Verbs\;}
            & \multicolumn{4}{c}{Ground-Truth Verb}  
        \\
        \hline
            &       &        & grnd
            &       &        & grnd 
            &       &        & grnd & grnd   
        \\
            Method
            & verb & value & value
            & verb & value & value
            & value & value-all & value & value-all
        \\
        \hline
        \hline
            w/o Gaze-S1 Transformer
            & 42.46 & 34.21 & 28.23
            & 70.89 & 55.47 & 45.34  
            & 76.02 & 41.96 & 61.21 & 23.15
        \\
            w/o Gaze-S2 Transformer
            & 43.02 & 31.24 & 23.27 
            & 71.17 & 51.70 & 36.59 
            & 69.68 & 32.94 & 48.44 & 13.05
        \\
            w/o Noun Classifiers on Gaze-S1 Transformer
            & 41.30 & 33.33  & 27.50 
            & 69.76 & 55.05  & 44.96 
            & 75.97 & 41.94 & \textbf{61.32} & \textbf{23.39}
        \\
            w/o Gradient Flow from Gaze-S2 Transformer to Glance Transformer
            & 42.96 & 33.82 & 25.77
            & 70.97 & 54.59 & 41.11
            & 73.91 & 38.59 & 55.10 & 17.10
        \\
            w/o Verb Token in Gaze-S2 Transformer
            & 44.36 & 35.57 & 29.16 
            & 72.84 & 56.79 & 46.19
            & 74.53 & 39.83 & 60.07 & 21.83
        \\
            \cellcolor[gray]{0.9}\mbox{CoFormer} (Ours)
            & \cellcolor[gray]{0.9}\textbf{44.41} & \cellcolor[gray]{0.9}\textbf{35.87} & \cellcolor[gray]{0.9}\textbf{29.37} 
            & \cellcolor[gray]{0.9}\textbf{72.98} & \cellcolor[gray]{0.9}\textbf{57.58} & \cellcolor[gray]{0.9}\textbf{46.70} 
            & \cellcolor[gray]{0.9}\textbf{76.17} & \cellcolor[gray]{0.9}\textbf{42.11} & \cellcolor[gray]{0.9}61.15 & \cellcolor[gray]{0.9}23.09
        \\
        \hline
    \end{tabular}}
    \vspace{-1mm}
    \caption{
        Ablation study of \mbox{CoFormer} on the SWiG dev set.
        The contributions of different components used in our model are evaluated.
    }
    \label{table:ablation}
\end{table*}

\mbox{CoFormer} is evaluated on the SWiG dataset~\cite{pratt2020grounded}, which is constructed by adding box annotations to the imSitu dataset~\cite{yatskar2016situation}.
The imSitu dataset contains 75K, 25K and 25K images for train, development and test set, respectively.
This dataset contains 504 verbs, 11K nouns and 190 roles.
The number of roles in the frame of a verb ranges from 1 to 6. 
Each image is paired with the annotation of a verb, and three nouns from three different annotators for each role.
In addition to this annotation, the SWiG dataset provides a box annotation for each role (except role \textit{Place}).

\subsection{Evaluation Metric}
\noindent \textbf{Metric Details.} 
The prediction accuracy of verb is measured by \textit{verb}, that of noun is evaluated by \textit{value} and \mbox{\textit{value-all}}, and that of grounded noun is assessed by \mbox{\textit{grounded-value}} and \mbox{\textit{grounded-value-all}}.
Regarding to the noun metrics, \textit{value} measures whether a noun is correct for each role, and \mbox{\textit{value-all}} measures whether all nouns are correct for entire roles in a frame simultaneously. 
The noun prediction is considered correct if the predicted noun matches any of the three noun annotations given by three annotators.
For the grounded noun metrics, \mbox{\textit{grounded-value}} measures whether a noun and its grounding are correct for each role, and \mbox{\textit{grounded-value-all}} measures whether all nouns and their groundings are correct for entire roles in a frame simultaneously. 
The grounding prediction is considered correct if the predicted box existence is correct and the predicted bounding box has \mbox{Intersection-over-Union} (IoU) value at least 0.5 with the box annotation.
Note that the above metrics are calculated per verb and then averaged over all verbs, since the number of roles in a frame depends on a verb and each verb might be associated with a different number of samples in the dataset.
\begin{figure*}[t!]
    \centering
    \includegraphics[width=\textwidth]{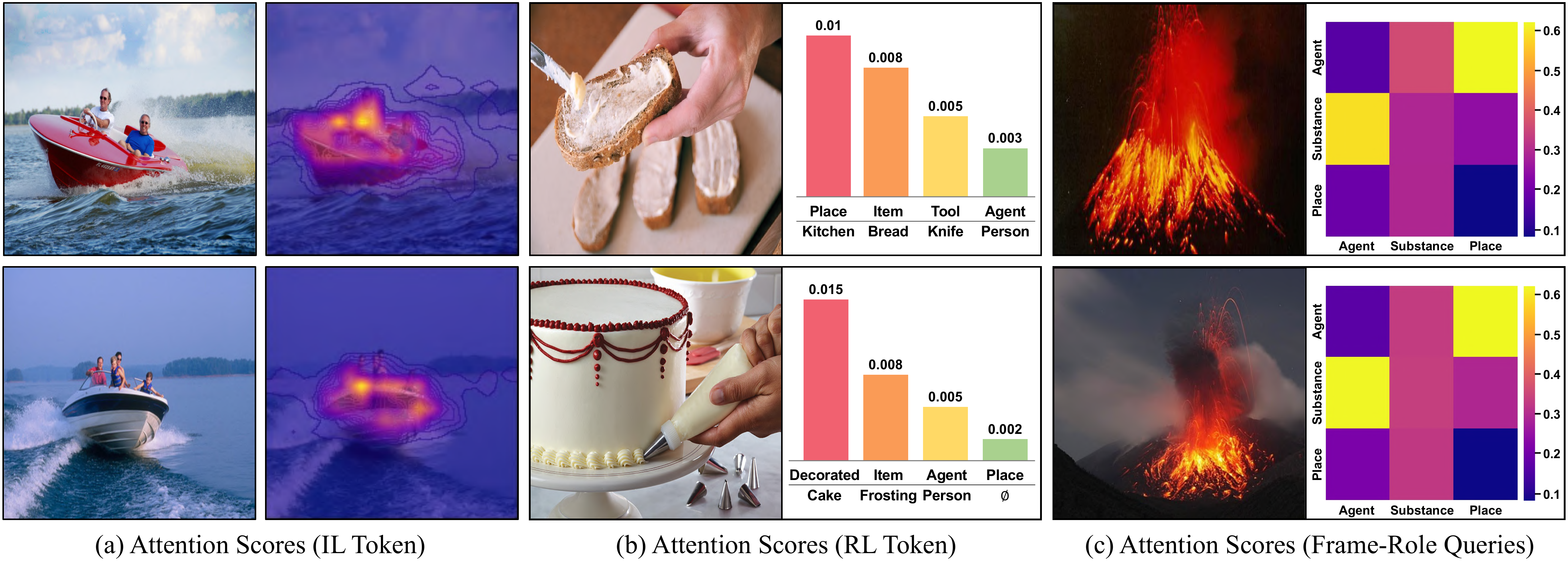}
    \caption{
    Attention scores from IL token to image features, from RL token to role features, and on \mbox{frame-role} queries.
    We visualize the attention scores computed from the last \mbox{self-attention} layer of the encoder in Glance transformer, the encoder in \mbox{Gaze-S1} transformer, and the decoder in \mbox{Gaze-S2} transformer, respectively.
    Higher attention scores are highlighted in red color on images.
    }
    \label{fig:qual_verb}
\end{figure*}
\begin{figure*}[t!]
    \centering
    \includegraphics[width=\textwidth]{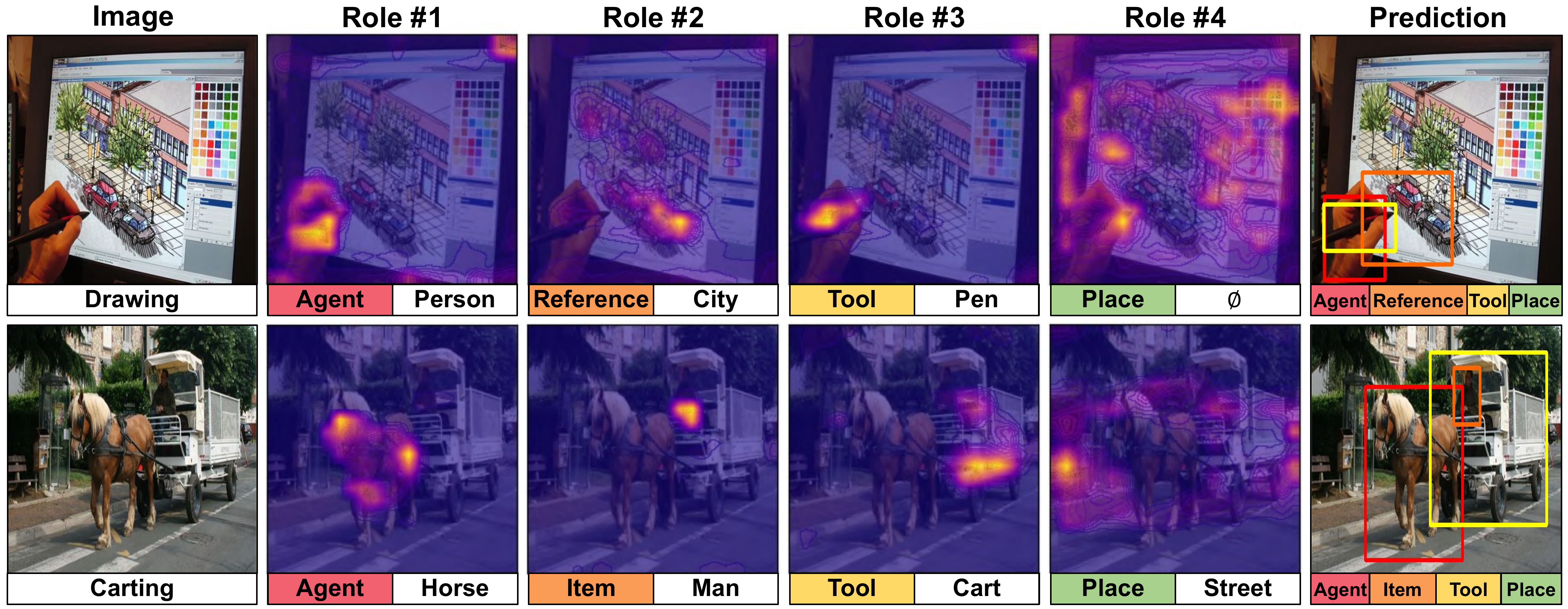}
    \caption{
    Attentions scores from \mbox{frame-role} queries to image features.
    We visualize the attention scores computed from the last cross-attention layer of the decoder in \mbox{Gaze-S2} transformer.
    Higher attention scores are highlighted in red color on images.
    }
    \label{fig:qual_role_attn}
\end{figure*}

\noindent \textbf{Evaluation Settings.} 
Three evaluation settings are proposed for comprehensive evaluation: \textit{\mbox{Top-1} Predicted Verb}, \textit{\mbox{Top-5} Predicted Verbs}, and \textit{\mbox{Ground-Truth} Verb}.
In \textit{\mbox{Top-1} Predicted Verb} setting, the predicted nouns and their groundings are considered incorrect if the \mbox{top-1} verb prediction is incorrect.
In \textit{\mbox{Top-5} Predicted Verbs} setting, the predicted nouns and their groundings are considered incorrect if the \mbox{ground-truth} verb is not contained in the \mbox{top-5} predicted verbs.
In \textit{\mbox{Ground-Truth} Verb} setting, the predicted nouns and their groundings are obtained by conditioning on the \mbox{ground-truth} verb.
\begin{figure*}[t!]
    \centering
    \includegraphics[width=0.965\textwidth]{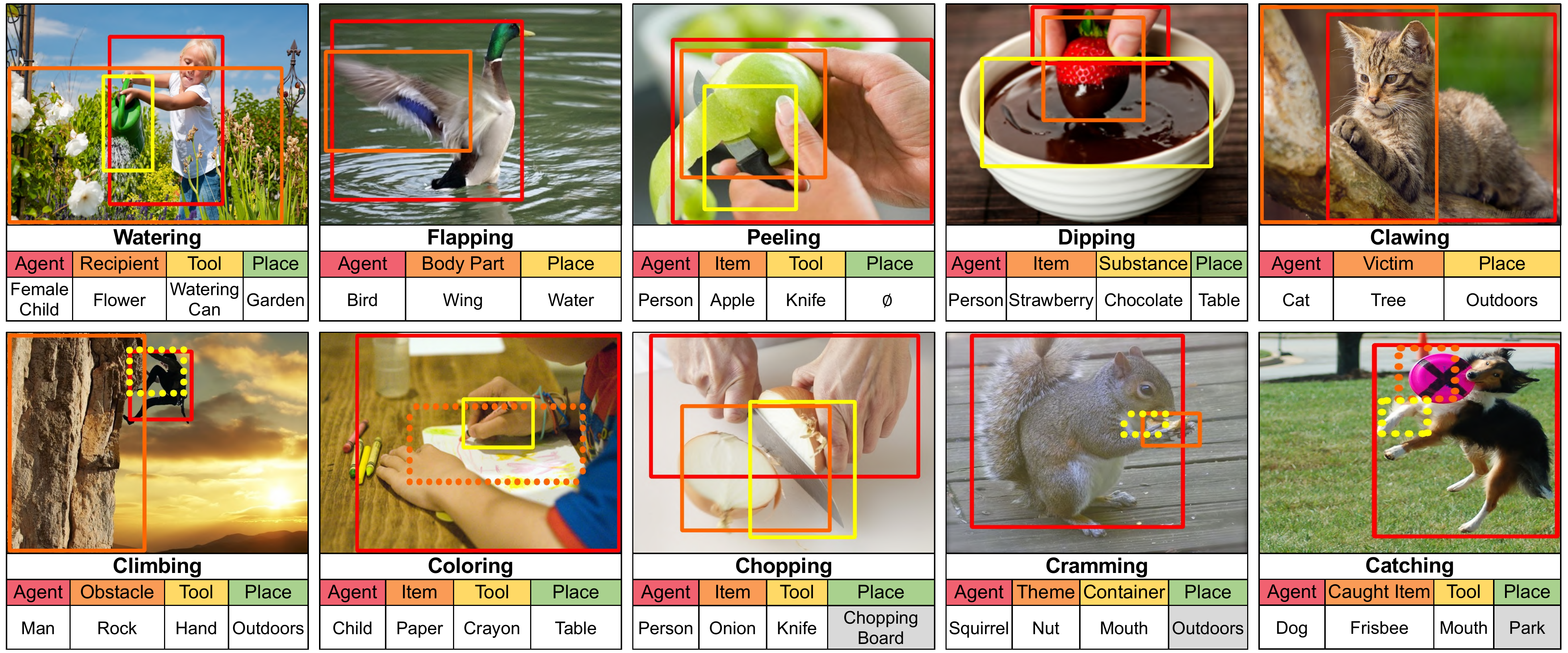}
    \caption{
    Prediction results.
    Dashed boxes denote incorrect grounding predictions.
    Incorrect noun predictions are highlighted in gray color.
    }
    \vspace{-2mm}
    \label{fig:qual_pred}
\end{figure*}

\subsection{Implementation Details}
We use \mbox{ResNet-50}~\cite{resnet} pretrained on ImageNet~\cite{deng2009imagenet} as a CNN backbone following existing models~\cite{pratt2020grounded,cho2021gsrtr} in GSR.
Given an image, the CNN backbone extracts image features of size $h \times w \times c$, where $h = w = 22$ and $c=2048$.
The embedding dimension of each token is $d=512$.
We employ AdamW Optimizer~\cite{loshchilov2018decoupled} with $10^{-4}$ weight decay, $\beta_1=0.9$, and $\beta_2=0.999$.
We train \mbox{CoFormer} with $10^{-4}$ learning rate ($10^{-5}$ for the CNN backbone) which decreases by a factor of $10$ at epoch $30$.
Training \mbox{CoFormer} with batch size of $16$ for $40$ epochs takes about $30$ hours on four RTX 3090 GPUs.
Complete details including loss coefficients are provided in the supplementary material.

\subsection{Quantitative Evaluations}
\mbox{CoFormer} achieves the state of the art in all evaluations as shown in Table~\ref{table:result}.
Existing SR models~\cite{mallya2017recurrent, li2017situation, suhail2019mixture, cooray2020attention} use at least two \mbox{VGG-16}~\cite{vggnet} backbones, and GSR models~\cite{pratt2020grounded} employ two \mbox{ResNet-50}~\cite{resnet} backbones for verb and noun prediction, while \mbox{CoFormer} only employs a single \mbox{ResNet-50} backbone. 
Compared with GSRTR~\cite{cho2021gsrtr},
the improvements in the verb prediction accuracies range from 3.35\%p to 4.03\%p.
Regarding to the noun prediction accuracies, 
the improvements range from 1.84\%p to 3.89\%p, and those in the grounded noun prediction accuracies range from 2.11\%p to 4.09\%p.
These results demonstrate that the proposed
collaborative framework is effective for GSR.

\noindent \textbf{Ablation Study.} 
We analyze the effects of different components in \mbox{CoFormer} as shown in Table~\ref{table:ablation}.
When we train our model without using \mbox{Gaze-S1} transformer or \mbox{Gaze-S2} transformer,
the accuracies in verb prediction or grounded noun prediction 
largely decrease, which demonstrates 
the effectiveness of the collaborative framework.
Training our \mbox{CoFormer} without using the two noun classifiers placed on top of \mbox{Gaze-S1} transformer leads to significant drops in the verb prediction accuracies.
In this case, it is difficult for role features to learn involved nouns and their relations, while the encoder in \mbox{Gaze-S1} transformer aggregates the role features through self-attentions.
To figure out whether \mbox{Gaze-S2} transformer assists Glance transformer by forcing it to implicitly consider involved nouns, we train \mbox{CoFormer} by restricting the flow of loss gradients through the aggregated image features from \mbox{Gaze-S2} transformer to Glance transformer.
As shown in the fourth row of Table~\ref{table:ablation}, the verb prediction accuracies drop, which demonstrates that \mbox{Gaze-S2} transformer supports Glance transformer via loss gradients through the aggregated image features. 
In \mbox{CoFormer}, each \mbox{frame-role} query is constructed by an addition of a role token embedding and a verb token embedding.
We study how effective it is by training \mbox{CoFormer} without using a verb token embedding for the construction of \mbox{frame-role} queries. 
The fifth row of Table~\ref{table:ablation} shows that the grounded noun prediction accuracies drop, 
which demonstrates that the verb token embedding is helpful for grounded noun prediction.

\subsection{Qualitative Evaluations}
We visualize the attention scores computed in the attention layers of \mbox{CoFormer}.  
Figure~\ref{fig:qual_verb}(a) shows that IL token captures the essential features to estimate a verb for two \textit{Boating} images.
Figure~\ref{fig:qual_verb}(b) shows how much RL token focuses on the roles in the frame of the \mbox{ground-truth} verb, and the classification results from the noun classifier placed on top of the encoder in \mbox{Gaze-S1} transformer; attention scores among 190 roles sum to 1.
This demonstrates that RL token effectively
captures involved nouns and their relations through self-attentions in the encoder of \mbox{Gaze-S1} transformer. 
Figure~\ref{fig:qual_verb}(c) shows how role relations are captured through self-attentions on \mbox{frame-role} queries, which demonstrates that \mbox{CoFormer} similarly captures the relations if the situations in images are similar; attention scores sum to 1 in each column. 
Figure~\ref{fig:qual_role_attn} shows the local regions where \mbox{frame-role} queries focus on, and the predicted grounded nouns corresponding to the queries.
Figure~\ref{fig:qual_pred} shows prediction results of \mbox{CoFormer} on the SWiG test set.
The first row shows the correct predictions, and the second row shows several incorrect predictions.

%% file: arxiv_5_conclusion.tex
\section{Conclusion}
We propose a collaborative framework for GSR, where the two processes for verb prediction and noun estimation interact and complement each other.
Using this framework, we present \mbox{CoFormer} which outperforms existing methods in all evaluation metrics on the SWiG dataset.
We also provide in-depth analyses of how \mbox{CoFormer} draws attentions on images and captures role relations with the ablation study on the effects of different components used in our model.
A limitation of \mbox{CoFormer} is that the model sometimes suffers from predicting the boxes which have extreme aspect ratios or small scales. 
This issue will be explored in future work.

%% file: arxiv_6_acknowledgement.tex
\vspace{1mm}
{\small
\noindent \textbf{Acknowledgement.} 
This work was supported by 
the NRF grant and 
the IITP grant 
funded by Ministry of Science and ICT, Korea
(NRF-2021R1A2C3012728, 
 No.2019-0-01906 Artificial Intelligence Graduate School Program--POSTECH, 
 No.2021-0-02068 Artificial Intelligence Innovation Hub, 
 IITP-2020-0-00842). 
}

%% file: arxiv_supp_0_intro.tex
This supplementary material provides 
method details 
(Section~\ref{supp:method}), implementation details (Section~\ref{supp:imple_details}), qualitative evaluations (Section~\ref{supp:qualitative}), an application of this task (Section~\ref{supp:application}), computational evaluations (Section~\ref{supp:computational_evaluations}) and a limitation (Section~\ref{supp:limitation}),
which could not be included
in the main paper due to the limited space.

%% file: arxiv_supp_1_method.tex
\section{Method Details}
\renewcommand\thefigure{A\arabic{figure}}
\label{supp:method}
Transformer architectures in our \mbox{CoFormer} consist of common building blocks, encoder and decoder.
The main components of these building blocks are attention layers.
Section~\ref{supp:attn_layers} provides more details of the attention layers. 

In Section 3.5 of the main paper, the losses to train our model are described: verb classification loss, noun classification losses, box existence prediction loss, and box regression losses.
Section~\ref{supp:loss_details} provides more details of the losses.

\subsection{Attention Layer}
\label{supp:attn_layers}

\noindent \textbf{Multi-Head Attention.}
The input of the multi-head attention layer is the sequence of query, key and value.
The query sequence is denoted by $\mathbf Q\in\mathbb R^{L_Q \times d}$, where $L_Q$ is the sequence length and $d$ is the size of the hidden dimension.
The key sequence is denoted by $\mathbf K \in\mathbb R^{L_{KV} \times d}$ and value sequence is denoted by $\mathbf V \in\mathbb R^{L_{KV} \times d}$, where $L_{KV}$ is the sequence length.
In the multi-head attention layer, we employ $H$ attention heads; the hidden dimension of each attention head is $d_h=d/H$.
For each attention head $i$, $\mathbf Q$, $\mathbf K$ and $\mathbf V$ are 
linearly 
projected via parameter matrices $\mathbf W^{Q}_i, \mathbf W^{K}_i, \mathbf W^{V}_i\in\mathbb R^{d\times d_h}$.
In details,
\begin{align}
    \mathbf Q_i &= \mathbf Q \mathbf W^{Q}_i \in \mathbb R^{L_Q \times d_h}\tag{A.1},\\
    \mathbf K_i &= \mathbf K \mathbf W^{K}_i \in \mathbb R^{L_{KV} \times d_h}\tag{A.2},\\
    \mathbf V_i &= \mathbf V \mathbf W^{V}_i \in \mathbb R^{L_{KV} \times d_h}\tag{A.3}.
\end{align}
The output of each attention head $i$ is obtained by a weighted summation of the value $\mathbf V_i$, where the weights are computed by the scaled dot-product between the query $\mathbf Q_i$ and the key $\mathbf K_i$ followed by a softmax function.
In details,
\begin{align}\tag{A.4}
    \mathrm{Attention}(\mathbf Q_i, \mathbf K_i, \mathbf V_i) = \mathrm{Softmax}(\frac{\mathbf Q_i \mathbf K_i ^T}{\sqrt{d_h}}) \mathbf V_i.
\end{align}
The output of each attention head $i$ is concatenated along hidden dimension, then linearly projected via a parameter matrix $\mathbf{W}^O\in\mathbb R^{d \times d}$.
In details,
\begin{align}\tag{A.5}
    \mathrm{MultiHead}(\mathbf Q, \mathbf K, \mathbf V) =  [\mathrm{Head}_1;...;\mathrm{Head}_H] \mathbf W^O,
\end{align}
where $[;]$ is a concatenation along hidden dimension and $\mathrm{Head}_i=\mathrm {Attention}(\mathbf Q_i, \mathbf K_i, \mathbf V_i)$ for $i=1,...,H$.

\noindent \textbf{Multi-Head Cross-Attention.}
This is the multi-head attention layer where the key sequence $\mathbf K$ is same with the value sequence $\mathbf V$, but the query sequence $\mathbf Q$ is different.

\noindent \textbf{Multi-Head Self-Attention.}
This is the multi-head attention layer where the query sequence $\mathbf Q$, key sequence $\mathbf K$, and value sequence $\mathbf V$ are same, \ie, $\mathbf Q = \mathbf K = \mathbf V$.

\subsection{Loss}
\label{supp:loss_details}
\begin{figure*}[t!]
    \centering
    \includegraphics[width=\textwidth]{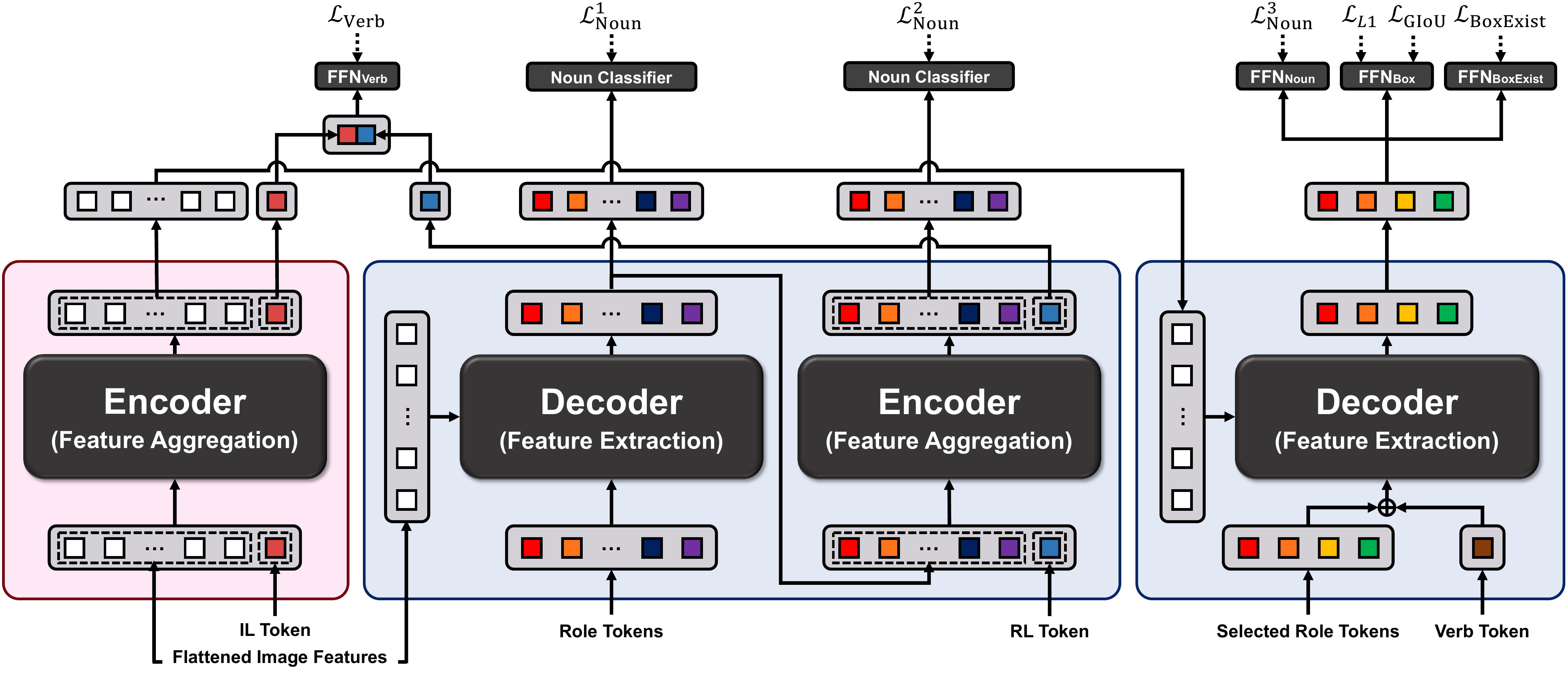}
    \caption{
    Transformer architectures in \mbox{CoFormer} including the losses to train our model. 
    The losses for training our \mbox{CoFormer} are as follows: 
    $\mathcal L_{\mathrm {Verb}},
    \mathcal L^1_{\mathrm {Noun}},
    \mathcal L^2_{\mathrm {Noun}},
    \mathcal L^3_{\mathrm {Noun}},
    \mathcal L_{\mathrm {Box Exist}}, 
    \mathcal L_{L1},
    \mathcal L_{\mathrm {GIoU}}$.
    }
    \label{fig:supp_model_loss}
\end{figure*}

Figure~\ref{fig:supp_model_loss} shows the losses to train \mbox{CoFormer}. 
The verb classification loss is denoted by $\mathcal L_{\mathrm {Verb}}$.
The noun classification loss from the classifier involved in the decoder of \mbox{Gaze-S1} transformer is denoted by $\mathcal L^1_{\mathrm {Noun}}$, the loss from the classifier involved in the encoder of \mbox{Gaze-S1} transformer is denoted by $\mathcal L^2_{\mathrm {Noun}}$, and the loss from the classifier involved in the decoder of \mbox{Gaze-S2} transformer is denoted by $\mathcal L^3_{\mathrm {Noun}}$.
The box existence prediction loss is denoted by $\mathcal L_{\mathrm {Box Exist}}$.
The $L1$ box regression loss is denoted by $\mathcal L_{L1}$.
The $\mathrm {GIoU}$ box regression loss is denoted by $\mathcal L_{\mathrm {GIoU}}$.

The total training loss is the linear combination of $\mathcal L_{\mathrm {Verb}},
\mathcal L^1_{\mathrm {Noun}},
\mathcal L^2_{\mathrm {Noun}},
\mathcal L^3_{\mathrm {Noun}},
\mathcal L_{\mathrm {Box Exist}}, 
\mathcal L_{L1},$ and
$\mathcal L_{\mathrm {GIoU}}$.
In this total loss, the loss coefficients are as follows:
$\lambda_{\mathrm {Verb}},
\lambda^1_{\mathrm {Noun}},
\lambda^2_{\mathrm {Noun}},
\lambda^3_{\mathrm {Noun}},
\lambda_{\mathrm {Box Exist}},
\lambda_{L1},
\lambda_{\mathrm {GIoU}} > 0$.

%% file: arxiv_supp_2_implementation.tex
\section{Implementation Details}
\renewcommand\thefigure{B\arabic{figure}}
\label{supp:imple_details} 
In Section 4.2 of the main paper, \mbox{some implementation} details are described.
For completeness, we describe more architecture details (Section~\ref{supp:imple_arch}), loss details (Section~\ref{supp:imple_loss}), augmentation details (Section~\ref{supp:imple_aug}), and training details (Section~\ref{supp:imple_train}) of our \mbox{CoFormer}. 

\subsection{Architecture Details}
\label{supp:imple_arch}
Following previous work~\cite{pratt2020grounded,cho2021gsrtr}, we use \mbox{ResNet-50}~\cite{resnet} pretrained on ImageNet~\cite{deng2009imagenet} as a CNN backbone.
Given an image, the CNN backbone produces image features of size $h \times w \times c$, where $h = w = 22$ and $c=2048$.
A $1 \times 1$ convolution followed by a flatten operation produces flattened image features $\mathbf X_{F} \in \mathbb R^{hw \times d}$, where $d=512$.
To retain spatial information, we employ positional encodings.
We use learnable 2D embeddings for the positional encodings.

We initialize encoders and decoders using Xavier Initialization~\cite{xavier2010init}, and these modules are trained with the dropout rate of $0.15$.
The number of heads in the attention layers of these modules is $8$.
Each of feed forward networks in these modules is \mbox{$2$-fully} connected layers with a ReLU activation function, whose hidden dimensions are $4d$ and dropout rate is $0.15$.
These modules take learnable tokens, and each embedding dimension of the tokens is $d$.

The verb classifier $\mathrm{FFN_{\mathrm {Verb}}}$ is \mbox{$2$-fully} connected layers with a ReLU activation function, whose hidden dimensions are $2d$ and dropout rate is $0.3$.
Each of the two noun classifiers placed on top of \mbox{Gaze-S1} transformer is
a linear layer.
The noun classifier $\mathrm{FFN_{\mathrm {Noun}}}$ is \mbox{$2$-fully} connected layers with a ReLU activation function, whose hidden dimensions are $2d$ and dropout rate is $0.3$.
The bounding box estimator $\mathrm{FFN_{\mathrm {Box}}}$ is \mbox{$3$-fully} connected layers with two ReLU activation functions, whose hidden dimensions are $2d$ and dropout rate is $0.2$.
The box existence predictor $\mathrm{FFN_{\mathrm {Box Exist}}}$ is \mbox{$2$-fully} connected layers with a ReLU activation function, whose hidden dimensions are $2d$ and dropout rate is $0.2$.

\subsection{Loss Details}
\label{supp:imple_loss}

\noindent \textbf{Complete Details of Noun Losses.}
In the SWiG dataset, each image is associated with three noun annotations given by three different annotators for each role.
For the noun classification losses $\mathcal L^1_{\mathrm {Noun}}$, $\mathcal L^2_{\mathrm {Noun}}$, $\mathcal L^3_{\mathrm {Noun}}$, each noun loss is obtained by the summation of three classification losses 
corresponding to three different annotators.

\noindent \textbf{Regularization.}
We employ label smoothing regularization~\cite{szegedy2016rethinking} in the loss computation for verb classification loss $\mathcal L_{\mathrm {Verb}}$ and noun classification losses $\mathcal L^1_{\mathrm {Noun}}$, $\mathcal L^2_{\mathrm {Noun}}$, $\mathcal L^3_{\mathrm {Noun}}$.
In details, the label smoothing factor in the computation of verb classification loss is $0.3$, and the factor in the computation of noun classification losses is $0.2$.

\noindent \textbf{Loss Coefficients.}
Total loss to train \mbox{CoFormer} is a linear combination of losses.
In our implementation, the loss coefficients are $\lambda_{\mathrm {Verb}}=\lambda^3_{\mathrm {Noun}}=1$, $\lambda^1_{\mathrm {Noun}}=\lambda^2_{\mathrm {Noun}}=2$, and
$\lambda_{\mathrm {Box Exist}}=\lambda_{L1}= \lambda_{\mathrm {GIoU}}=5$.

\subsection{Augmentation Details}
\label{supp:imple_aug}
For data augmentation, we employ random scaling, random horizontal flipping, random color jittering, and random gray scaling.
The input images are randomly scaled with the scaling factors of $0.5$, $0.75$, and $1.0$.
Also, the input images are horizontally flipped with the probability of $0.5$.
The brightness, saturation and hue of the input images are randomly changed with the factor of $0.1$ for each change.
The input images are randomly converted to grayscale with the probability of $0.3$.

\renewcommand\thefigure{C\arabic{figure}}
\begin{figure*}[t!]
    \centering
    \includegraphics[width=\textwidth]{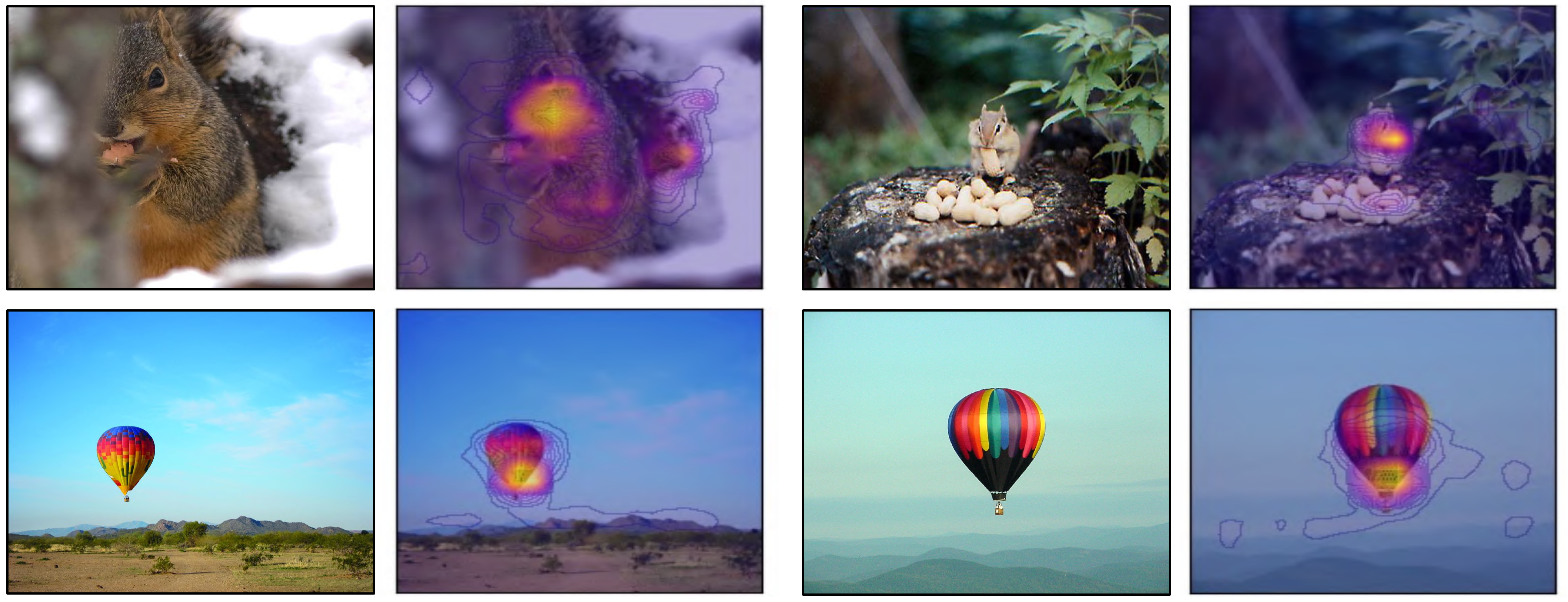}
    \caption{
    Attention scores from IL token to image features.
    We visualize the attention scores computed from the last \mbox{self-attention} layer of the encoder in Glance transformer.
    Higher attention scores are highlighted in red color on images.
    }
    \label{fig:supp_il_tok}
    \vspace{0.5mm}
\end{figure*}
\begin{figure*}[t!]
    \centering
    \includegraphics[width=\textwidth]{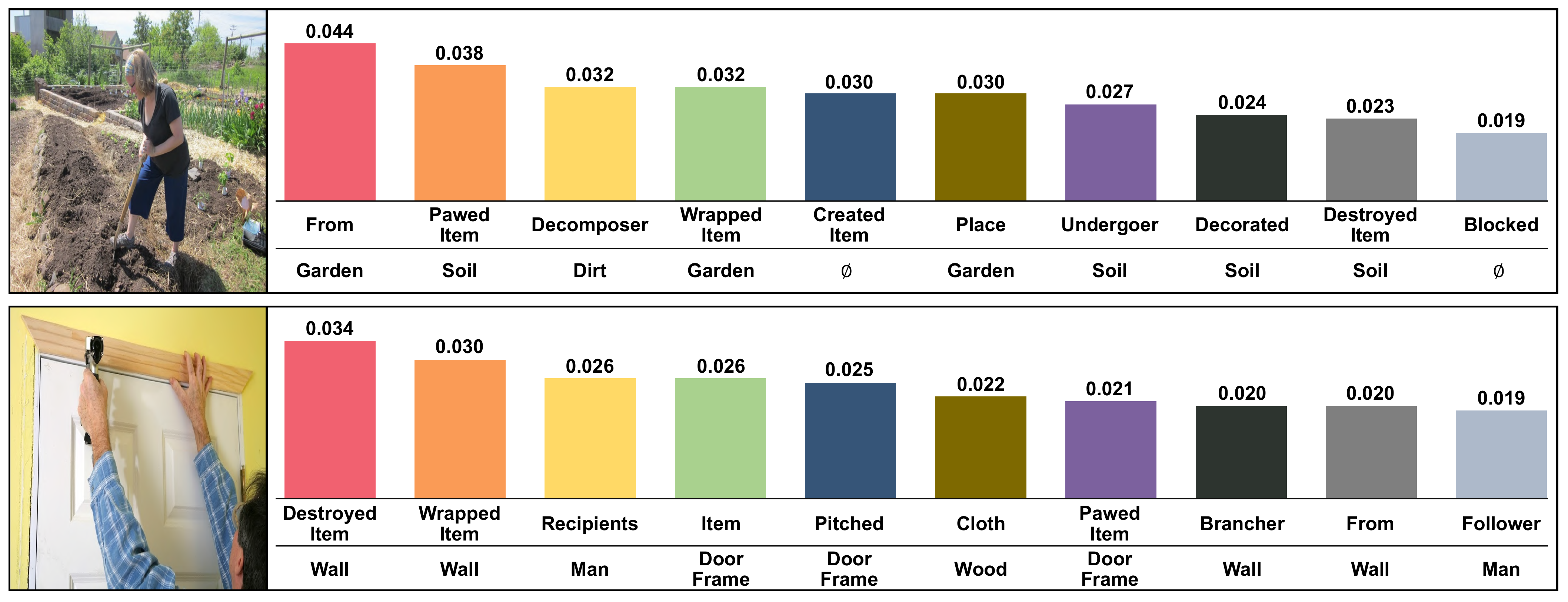}
    \caption{
    Attention scores from RL token to role features.
    We visualize the attention scores computed from the last \mbox{self-attention} layer of the encoder in \mbox{Gaze-S1} transformer.
    Note that we show the roles where RL token has top-10 attentions scores.
    In Figure 7(b) of the main paper, we show the results corresponding to the roles in the frame of the ground-truth verb.
    }
    \label{fig:supp_rl_tok}
    \vspace{0.5mm}
\end{figure*}

\subsection{Training Details}
\label{supp:imple_train}
We employ AdamW Optimizer~\cite{loshchilov2018decoupled} with the weight decay of $10^{-4}$, $\beta_1=0.9$, and $\beta_2=0.999$.
For stable training, we apply gradient clipping with the maximal gradient norm of $0.1$.
The transformers, classifiers and learnable embeddings are trained with the learning rate of $10^{-4}$.
The CNN backbone is \mbox{fine-tuned} with the learning rate of $10^{-5}$.
Note that we have a learning rate scheduler and the learning rates are divided by $10$ at epoch 30.
For batch training, we set the batch size to $16$.
We train \mbox{CoFormer} for $40$ epochs, which takes about $30$ hours on four RTX 3090 GPUs.

\newpage

%% file: arxiv_supp_3_qualitative.tex
\section{Qualitative Evaluations}
\renewcommand\thefigure{C\arabic{figure}}
\label{supp:qualitative}
\begin{figure*}[t!]
    \centering
    \includegraphics[width=\textwidth]{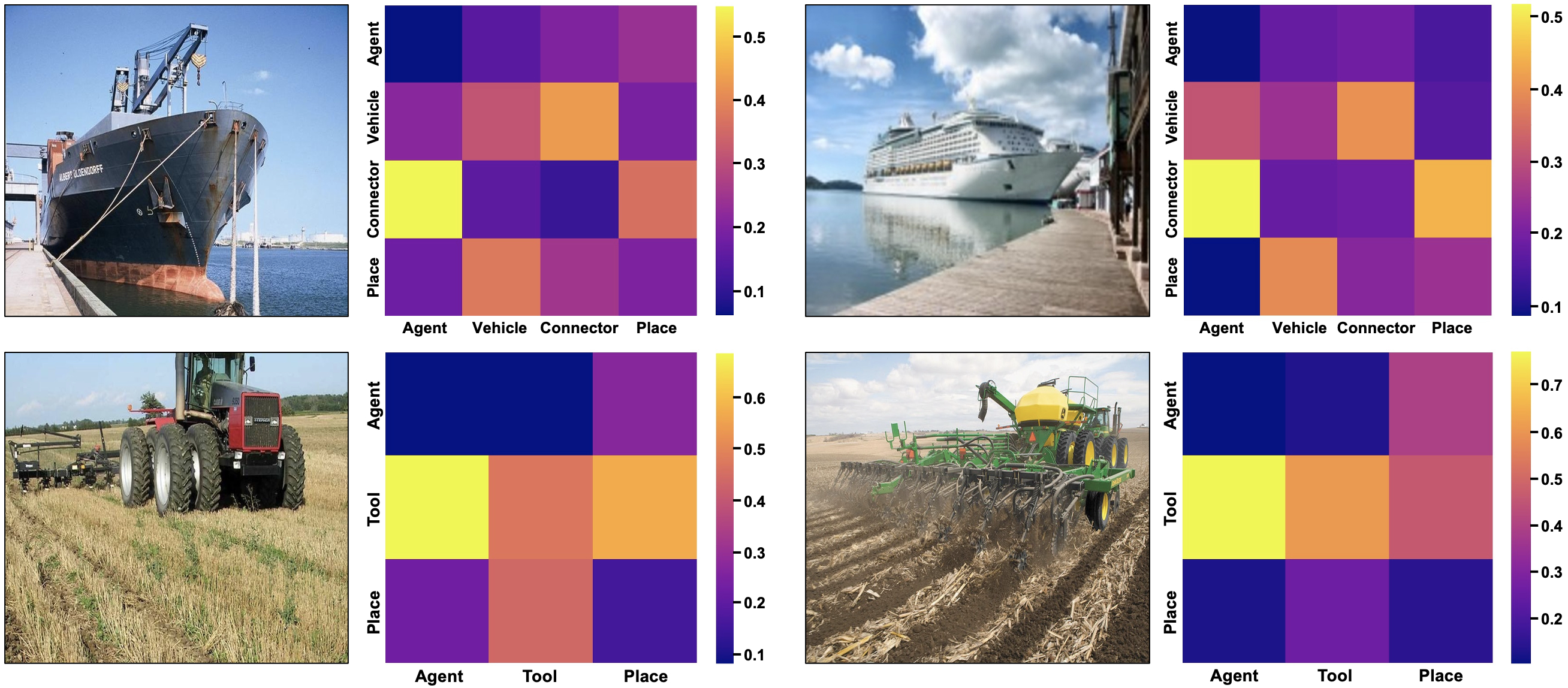}
    \caption{
    Attention scores on \mbox{frame-role} queries.
    We visualize the attention scores computed from the last \mbox{self-attention} layer of the decoder in \mbox{Gaze-S2} transformer.
    }
    \label{fig:supp_relations}
    \vspace{0.5mm}
\end{figure*}
\begin{figure*}[t!]
    \centering
    \includegraphics[width=\textwidth]{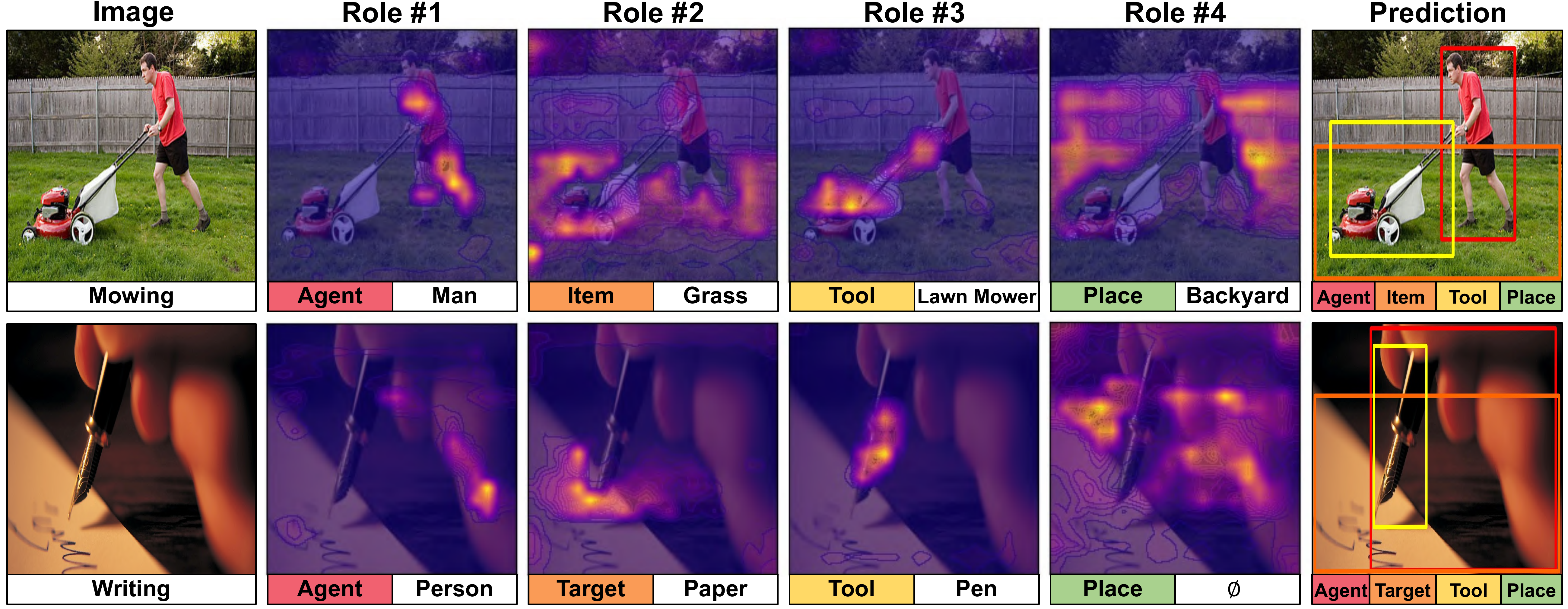}
    \caption{
    Attentions scores from \mbox{frame-role} queries to image features.
    We visualize the attention scores computed from the last cross-attention layer of the decoder in \mbox{Gaze-S2} transformer.
    Higher attention scores are highlighted in red color on images.
    }
    \label{fig:supp_img_attn}
    \vspace{0.5mm}
\end{figure*}
We visualize the attention scores computed in the attention layers of the transformers in our \mbox{CoFormer}.
Figure~\ref{fig:supp_il_tok} shows that IL token captures the essential features to estimate the main activities for two \textit{Cramming} images and two \textit{Ballooning} images.
Figure~\ref{fig:supp_rl_tok} shows the roles where RL token has top-10 attention scores, and the classification results from the noun classifier placed on top of the encoder in \mbox{Gaze-S1} transformer; attention scores among 190 roles sum to 1.
Note that several roles where RL token has high attention scores are not relevant to the main activity, 
but the noun classification results corresponding to those roles are highly relevant to the activity.
Since RL token leverages the role features which are fed as input to the noun classifier, it is reasonable to aggregate those role features for accurate verb prediction; the role features are aware of involved nouns and their relations.
Figure~\ref{fig:supp_rl_tok} demonstrates that RL token can effectively
capture involved nouns and their relations for \mbox{noun-aware} verb prediction
through self-attentions on the role features in the encoder of \mbox{Gaze-S1} transformer. 
Figure~\ref{fig:supp_relations} shows how role relations are captured through self-attentions on \mbox{frame-role} queries, which demonstrates that \mbox{CoFormer} similarly captures the role relations if the situations in images are similar; attention scores sum to 1 in each column. 
Figure~\ref{fig:supp_img_attn} shows the local regions where \mbox{frame-role} queries focus on, and the predicted grounded nouns corresponding to the queries.
This demonstrates that each query effectively captures its relevant local regions through cross-attentions between the queries and image features in the decoder of \mbox{Gaze-S2} transformer.
Note that those queries are constructed by leveraging the predicted verb, which significantly reduces the number of role candidates handled in noun estimation;
\mbox{Gaze-S1} transformer considers all role candidates, but \mbox{Gaze-S2} transformer handles a few roles associated with the predicted verb.
Figure~\ref{fig:supp_pred} shows the prediction results of \mbox{CoFormer} on the SWiG test set.
The first and second row show correct prediction results.
The third and fourth row show incorrect prediction results.
As shown in Figure~\ref{fig:supp_pred}, three noun annotations are given for each role in the SWiG dataset.
Note that the noun prediction is considered correct if the predicted noun matches any of the three noun annotations.
The grounded noun prediction is considered correct if a noun, a bounding box, and box existence are correctly predicted for a role.

\begin{figure*}[t!]
    \centering
    \includegraphics[width=\textwidth]{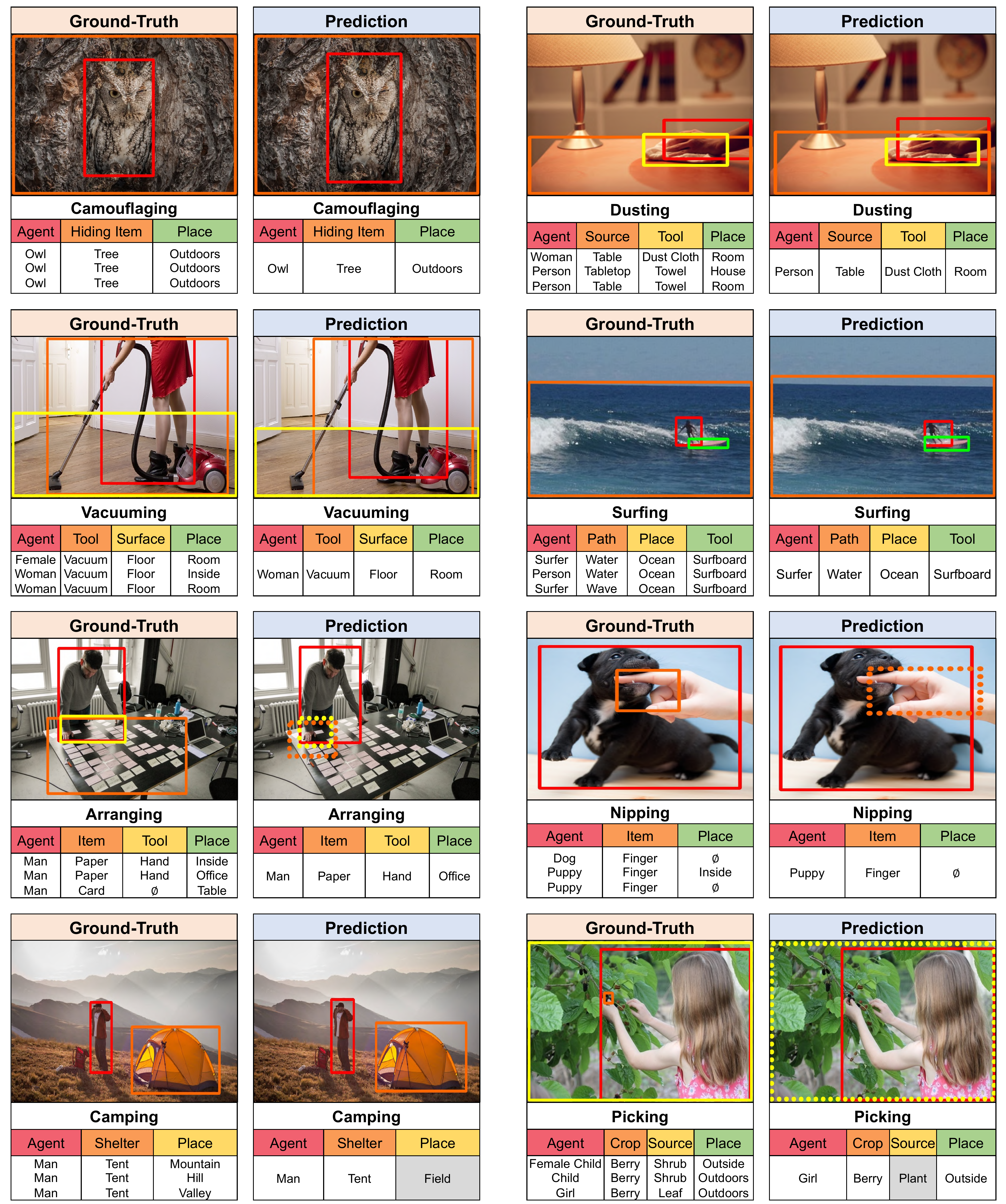}
    \caption{
    Prediction results of our \mbox{CoFormer} on the SWiG test set.
    Dashed boxes denote incorrect grounding predictions.
    Incorrect noun predictions are highlighted in gray color.
    }
    \label{fig:supp_pred}
\end{figure*}

%% file: arxiv_supp_4_application.tex
\renewcommand\thefigure{D\arabic{figure}}
\begin{figure*}[t!]
    \centering
    \includegraphics[width=\textwidth]{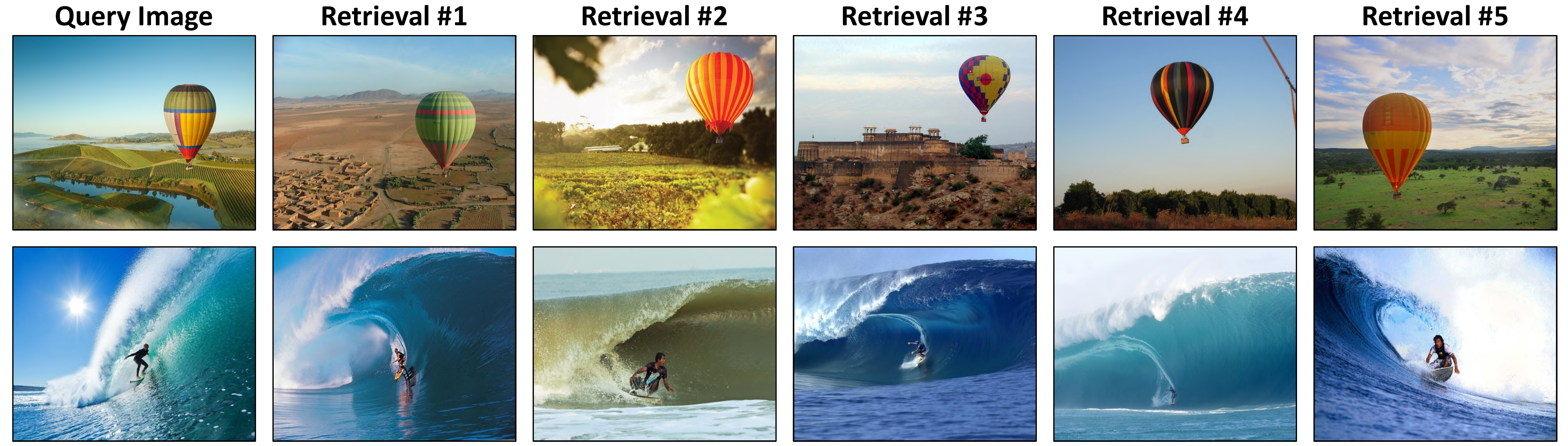}
    \caption{
    Grounded semantic aware image retrieval on the SWiG dev set.
    For each query image, we show the retrieval results which have top-5 similarity scores computed by $\mathrm{GrSitSim}(\cdot)$~\cite{pratt2020grounded}.
    This retrieval computes the similarity between two images considering 
    the predicted verbs, nouns, and \mbox{bounding-box} groundings of the nouns.
    }
    \label{fig:supp_retrieval}
\end{figure*}

\begin{figure*}[t!]
\begin{equation}\tag{D.6}
    \mathrm{GrSitSim}(I, J) = 
    \max \left\{ 
    \frac{\mathbbm 1_{[\hat{v}_i^{I} \; = \; \hat{v}_j^{J}]}}{i\cdot j \cdot \vert \mathcal R_{\hat{v}_i^I} \vert}
    \mathlarger{\sum_{k=1}^{\vert \mathcal{R}_{\hat{v}_{i}^{I}} \vert}}
    \mathbbm 1_{[\hat{n}_{i,k}^{I} \; = \; \hat{n}_{j,k}^{J}]}\cdot
    \left(1+\mathrm{IoU}(\mathbf{\hat{b'}}_{i,k}^{I}, \mathbf{\hat{b'}}_{j,k}^{J})\right)  
    \Bigg\vert  1 \leq i,j \leq 5\right\}.
    \label{eq:grsitsim}
\end{equation}
\end{figure*}

\section{Application}
\label{supp:application}
As shown in Figure~\ref{fig:supp_retrieval}, we can apply GSR models to grounded semantic aware image retrieval.
This image retrieval computes the similarity between two images considering their grounded situations.
In details, a similarity score between an image $I$ and an image $J$ is computed by $\mathrm{GrSitSim}(I, J)$~(Eq.~\ref{eq:grsitsim}).
Given an image $I$, a GSR model predicts the top-5 most probable verbs $\hat{v}_{1}^{I}, ..., \hat{v}_{5}^{I}$.
For each predicted verb $\hat{v}_{i}^{I}$, the model predicts nouns $\hat{n}_{i,1}^{I}, ..., \hat{n}_{i,|\mathcal{R}_{\hat{v}_{i}^{I}}|}^{I}$ and bounding boxes $\mathbf{\hat{b'}}_{i,1}^{I}$, ..., $\mathbf{\hat{b'}}_{i,|\mathcal{R}_{\hat{v}_{i}^{I}}|}^{I}$. 
These prediction results are used in the computation of $\mathrm{GrSitSim}(I, J)$.
By this score function, the similarity score is maximized if the top-1 predicted verb and the predicted grounded nouns are same for the two images $I$ and $J$.
Using this retrieval, we can retrieve images which have similar grounded situations with the situation of a query image.

\section{Computational Evaluations}
\label{supp:computational_evaluations}
\renewcommand\thetable{E\arabic{table}}
The number of parameters and inference time of our \mbox{CoFormer} are shown in Table~\ref{table:param}.
We also evaluate JSL~\cite{pratt2020grounded} and GSRTR~\cite{cho2021gsrtr} on the SWiG test set using a single 2080Ti GPU with a batch size of 1.
JSL uses two ResNet-50~\cite{resnet} and a feature pyramid network (FPN)~\cite{lin2017fpn} in the CNN backbone, while GSRTR and our \mbox{CoFormer} only employ a single ResNet-50 in the backbone;
these two models demand much shorter inference time than JSL, which is crucial for \mbox{real-world} applications.
GSRTR and \mbox{CoFormer} are trained in an \mbox{end-to-end} manner, but JSL is trained separately in terms of verb model and grounded noun model.

\begin{table}[!t]
    \centering
    \resizebox{\columnwidth}{!}{
        \begin{tabular}{l|ccc}
        \hline
        Method & Backbone & \#Params & Inference Time
        \\
        \hline
        \hline
            JSL~\cite{pratt2020grounded}
            & R50, R50-FPN & 108 M & 80.23 ms (12.46 FPS)
        \\
            GSRTR~\cite{cho2021gsrtr}
            & R50 & 83 M & 21.69 ms (46.10 FPS)
        \\
            \cellcolor[gray]{0.9}\mbox{CoFormer} (Ours)
            & \cellcolor[gray]{0.9}R50 & \cellcolor[gray]{0.9}93 M & \cellcolor[gray]{0.9}30.62 ms (32.66 FPS)
        \\
        \hline
    \end{tabular}}
    \caption{
        Number of parameters and inference time.
        Inference time was measured on the SWiG test set using one 2080Ti GPU.
    }
    \label{table:param}
\end{table}

\section{Limitation}
\label{supp:limitation}
\renewcommand\thetable{F\arabic{table}}
\begin{table}[!t]
    \centering
    \resizebox{\columnwidth}{!}{
        \begin{tabular}{l|ccc|ccc}
        \hline
        \multicolumn{1}{c|}{}
            & \multicolumn{3}{c|}{Area ($\mathrm{width} \times \mathrm{height}$)}
            & \multicolumn{3}{c}{Aspect Ratio ($\mathrm{width} /  \mathrm{height}$)}
        \\
        \hline
            Metric
            & 0-10\% & 10-20\% & 20-100\% 
            & 0-5\% & 5-95\% & 95-100\% 
        \\
        \hline
        \hline
            value
            & 66.82 & 69.68 & 78.64
            & 72.75 & 76.24 & 71.88
        \\
            grnd value
            & 7.42 & 25.38 & 65.49
            & 36.88 & 62.62 & 31.01
        \\
        \hline
    \end{tabular}}
    \caption{
        Quantitative analysis of our \mbox{CoFormer} on the SWiG dev set in Ground-Truth Verb evaluation setting.
        The effects of box scales and aspect ratios are evaluated.
        Each range denotes the ratio of ground-truth boxes when sorted by the value of area or aspect ratio in ascending order.
        }
    \label{table:supp_analysis}
\end{table}
As shown in Table~\ref{table:supp_analysis}, \mbox{CoFormer} suffers from estimating the noun labels and boxes for objects which have small scales (Area 0-10\% and 10-20\%) or extreme aspect ratios (Aspect Ratio 0-5\% and 95-100\%).
To overcome such limitation, one may leverage \mbox{multi-scale} image features.

%% file: arxiv_main.bbl
\begin{thebibliography}{10}\itemsep=-1pt

\bibitem{carion2020end}
Nicolas Carion, Francisco Massa, Gabriel Synnaeve, Nicolas Usunier, Alexander
  Kirillov, and Sergey Zagoruyko.
\newblock {End-to-End Object Detection with Transformers}.
\newblock In {\em Proceedings of the European Conference on Computer Vision
  (ECCV)}, pages 213--229, 2020.

\bibitem{chen2021human}
Long Chen, Zhihong Jiang, Jun Xiao, and Wei Liu.
\newblock {Human-Like Controllable Image Captioning With Verb-Specific Semantic
  Roles}.
\newblock In {\em Proceedings of the IEEE/CVF Conference on Computer Vision and
  Pattern Recognition (CVPR)}, pages 16846--16856, 2021.

\bibitem{cho2021gsrtr}
Junhyeong Cho, Youngseok Yoon, Hyeonjun Lee, and Suha Kwak.
\newblock {Grounded Situation Recognition with Transformers}.
\newblock In {\em Proceedings of the British Machine Vision Conference (BMVC)},
  2021.

\bibitem{cooray2020attention}
Thilini Cooray, Ngai-Man Cheung, and Wei Lu.
\newblock {Attention-Based Context Aware Reasoning for Situation Recognition}.
\newblock In {\em Proceedings of the IEEE/CVF Conference on Computer Vision and
  Pattern Recognition (CVPR)}, pages 4736--4745, 2020.

\bibitem{deng2009imagenet}
Jia Deng, Wei Dong, Richard Socher, Li-Jia Li, Kai Li, and Li Fei-Fei.
\newblock {ImageNet: A large-scale hierarchical image database}.
\newblock In {\em 2009 IEEE Conference on Computer Vision and Pattern
  Recognition}, pages 248--255, 2009.

\bibitem{dosovitskiy2021an}
Alexey Dosovitskiy, Lucas Beyer, Alexander Kolesnikov, Dirk Weissenborn,
  Xiaohua Zhai, Thomas Unterthiner, Mostafa Dehghani, Matthias Minderer, Georg
  Heigold, Sylvain Gelly, Jakob Uszkoreit, and Neil Houlsby.
\newblock {An Image is Worth 16x16 Words: Transformers for Image Recognition at
  Scale}.
\newblock In {\em International Conference on Learning Representations (ICLR)},
  2021.

\bibitem{fillmore2003background}
Charles~J. Fillmore, Christopher~R. Johnson, and Miriam~R.L. Petruck.
\newblock {Background to Framenet}.
\newblock {\em International Journal of Lexicography}, 16(3):235--250, 2003.

\bibitem{HabitFormation}
Benjamin Gardner and Amanda~L. Rebar.
\newblock {Habit Formation and Behavior Change}.
\newblock {\em Oxford research encyclopedia of psychology}, 2019.

\bibitem{xavier2010init}
Xavier Glorot and Yoshua Bengio.
\newblock {Understanding the difficulty of training deep feedforward neural
  networks}.
\newblock In {\em Proceedings of the Thirteenth International Conference on
  Artificial Intelligence and Statistics}, pages 249--256, 2010.

\bibitem{guo2020normalized}
Longteng Guo, Jing Liu, Xinxin Zhu, Peng Yao, Shichen Lu, and Hanqing Lu.
\newblock {Normalized and Geometry-Aware Self-Attention Network for Image
  Captioning}.
\newblock In {\em Proceedings of the IEEE/CVF Conference on Computer Vision and
  Pattern Recognition (CVPR)}, 2020.

\bibitem{resnet}
{He, Kaiming and Zhang, Xiangyu and Ren, Shaoqing and Sun, Jian}.
\newblock {Deep Residual Learning for Image Recognition}.
\newblock In {\em Proceedings of the IEEE Conference on Computer Vision and
  Pattern Recognition (CVPR)}, pages 770--778, 2016.

\bibitem{hochreiter1997long}
Sepp Hochreiter and Jürgen Schmidhuber.
\newblock {Long Short-Term Memory}.
\newblock {\em Neural Computation}, 9(8):1735--1780, 1997.

\bibitem{huang2019attention}
Lun Huang, Wenmin Wang, Jie Chen, and Xiao-Yong Wei.
\newblock {Attention on Attention for Image Captioning}.
\newblock In {\em Proceedings of the IEEE/CVF International Conference on
  Computer Vision (ICCV)}, pages 4634--4643, 2019.

\bibitem{kahneman2003maps}
Daniel Kahneman.
\newblock {Maps of Bounded Rationality: Psychology for Behavioral Economics}.
\newblock {\em The American Economic Review}, 93(5):1449--1475, 2003.

\bibitem{khandelwal2021segmentation}
Siddhesh Khandelwal, Mohammed Suhail, and Leonid Sigal.
\newblock {Segmentation-Grounded Scene Graph Generation}.
\newblock In {\em Proceedings of the IEEE/CVF International Conference on
  Computer Vision (ICCV)}, pages 15879--15889, 2021.

\bibitem{kim2021hotr}
Bumsoo Kim, Junhyun Lee, Jaewoo Kang, Eun-Sol Kim, and Hyunwoo~J. Kim.
\newblock {HOTR: End-to-End Human-Object Interaction Detection With
  Transformers}.
\newblock In {\em Proceedings of the IEEE/CVF Conference on Computer Vision and
  Pattern Recognition (CVPR)}, pages 74--83, 2021.

\bibitem{lafferty2001conditional}
John Lafferty, Andrew McCallum, and Fernando Pereira.
\newblock {Conditional Random Fields: Probabilistic Models for Segmenting and
  Labeling Sequence Data}.
\newblock In {\em International Conference on Machine Learning (ICML)}, pages
  282--289, 2001.

\bibitem{lee2021ctrl}
Jinwoo Lee, Hyunsung Go, Hyunjoon Lee, Sunghyun Cho, Minhyuk Sung, and Junho
  Kim.
\newblock {CTRL-C: Camera Calibration TRansformer With Line-Classification}.
\newblock In {\em Proceedings of the IEEE/CVF International Conference on
  Computer Vision (ICCV)}, pages 16228--16237, 2021.

\bibitem{li2017situation}
Ruiyu Li, Makarand Tapaswi, Renjie Liao, Jiaya Jia, Raquel Urtasun, and Sanja
  Fidler.
\newblock {Situation Recognition with Graph Neural Network}.
\newblock In {\em Proceedings of the IEEE International Conference on Computer
  Vision (ICCV)}, pages 4173--4182, 2017.

\bibitem{li2016gated}
Yujia Li, Daniel Tarlow, Marc Brockschmidt, and Richard Zemel.
\newblock {Gated Graph Sequence Neural Networks}.
\newblock In {\em International Conference on Learning Representations (ICLR)},
  2016.

\bibitem{li2019transferable}
Yong-Lu Li, Siyuan Zhou, Xijie Huang, Liang Xu, Ze Ma, Hao-Shu Fang, Yanfeng
  Wang, and Cewu Lu.
\newblock {Transferable Interactiveness Knowledge for Human-Object Interaction
  Detection}.
\newblock In {\em Proceedings of the IEEE/CVF Conference on Computer Vision and
  Pattern Recognition (CVPR)}, pages 3585--3594, 2019.

\bibitem{lin2017fpn}
Tsung-Yi Lin, Piotr Dollar, Ross Girshick, Kaiming He, Bharath Hariharan, and
  Serge Belongie.
\newblock {Feature Pyramid Networks for Object Detection}.
\newblock In {\em Proceedings of the IEEE Conference on Computer Vision and
  Pattern Recognition (CVPR)}, pages 2117--2125, 2017.

\bibitem{lin2017focal}
Tsung-Yi Lin, Priya Goyal, Ross Girshick, Kaiming He, and Piotr Doll{\'a}r.
\newblock {Focal Loss for Dense Object Detection}.
\newblock In {\em Proceedings of the IEEE International Conference on Computer
  Vision (ICCV)}, pages 2980--2988, 2017.

\bibitem{liu2021paint}
Songhua Liu, Tianwei Lin, Dongliang He, Fu Li, Ruifeng Deng, Xin Li, Errui
  Ding, and Hao Wang.
\newblock {Paint Transformer: Feed Forward Neural Painting With Stroke
  Prediction}.
\newblock In {\em Proceedings of the IEEE/CVF International Conference on
  Computer Vision (ICCV)}, pages 6598--6607, 2021.

\bibitem{loshchilov2018decoupled}
Ilya Loshchilov and Frank Hutter.
\newblock {Decoupled Weight Decay Regularization}.
\newblock In {\em International Conference on Learning Representations (ICLR)},
  2019.

\bibitem{lu2021context}
Yichao Lu, Himanshu Rai, Jason Chang, Boris Knyazev, Guangwei Yu, Shashank
  Shekhar, Graham~W. Taylor, and Maksims Volkovs.
\newblock {Context-Aware Scene Graph Generation With Seq2Seq Transformers}.
\newblock In {\em Proceedings of the IEEE/CVF International Conference on
  Computer Vision (ICCV)}, pages 15931--15941, 2021.

\bibitem{mallya2017recurrent}
Arun Mallya and Svetlana Lazebnik.
\newblock {Recurrent Models for Situation Recognition}.
\newblock In {\em Proceedings of the IEEE International Conference on Computer
  Vision (ICCV)}, pages 455--463, 2017.

\bibitem{Peters06_numeracy}
Ellen Peters, Daniel Västfjäll, Paul Slovic, C.K. Mertz, Ketti Mazzocco, and
  Stephan Dickert.
\newblock {Numeracy and Decision Making}.
\newblock {\em Psychological Science}, 17(5):407--413, 2006.

\bibitem{pratt2020grounded}
Sarah Pratt, Mark Yatskar, Luca Weihs, Ali Farhadi, and Aniruddha Kembhavi.
\newblock {Grounded Situation Recognition}.
\newblock In {\em Proceedings of the European Conference on Computer Vision
  (ECCV)}, pages 314--332, 2020.

\bibitem{rezatofighi2019generalized}
{Rezatofighi, Hamid and Tsoi, Nathan and Gwak, JunYoung and Sadeghian, Amir and
  Reid, Ian and Savarese, Silvio}.
\newblock {Generalized Intersection Over Union: A Metric and a Loss for
  Bounding Box Regression}.
\newblock In {\em Proceedings of the IEEE/CVF Conference on Computer Vision and
  Pattern Recognition (CVPR)}, pages 658--666, 2019.

\bibitem{vggnet}
{Simonyan, Karen and Zisserman, Andrew}.
\newblock Very deep convolutional networks for large-scale image recognition.
\newblock In {\em International Conference on Learning Representations (ICLR)},
  2015.

\bibitem{suhail2019mixture}
Mohammed Suhail and Leonid Sigal.
\newblock {Mixture-Kernel Graph Attention Network for Situation Recognition}.
\newblock In {\em Proceedings of the IEEE/CVF International Conference on
  Computer Vision (ICCV)}, pages 10363--10372, 2019.

\bibitem{szegedy2016rethinking}
Christian Szegedy, Vincent Vanhoucke, Sergey Ioffe, Jon Shlens, and Zbigniew
  Wojna.
\newblock {Rethinking the Inception Architecture for Computer Vision}.
\newblock In {\em Proceedings of the IEEE Conference on Computer Vision and
  Pattern Recognition (CVPR)}, pages 2818--2826, 2016.

\bibitem{vaswani2017attention}
Ashish Vaswani, Noam Shazeer, Niki Parmar, Jakob Uszkoreit, Llion Jones,
  Aidan~N Gomez, \L{}ukasz Kaiser, and Illia Polosukhin.
\newblock {Attention is All you Need}.
\newblock In {\em Advances in Neural Information Processing Systems (NIPS)},
  2017.

\bibitem{vinyals2015show}
Oriol Vinyals, Alexander Toshev, Samy Bengio, and Dumitru Erhan.
\newblock {Show and Tell: A Neural Image Caption Generator}.
\newblock In {\em Proceedings of the IEEE Conference on Computer Vision and
  Pattern Recognition (CVPR)}, 2015.

\bibitem{wang2020learning}
Tiancai Wang, Tong Yang, Martin Danelljan, Fahad~Shahbaz Khan, Xiangyu Zhang,
  and Jian Sun.
\newblock {Learning Human-Object Interaction Detection Using Interaction
  Points}.
\newblock In {\em Proceedings of the IEEE/CVF Conference on Computer Vision and
  Pattern Recognition (CVPR)}, 2020.

\bibitem{xiong2020layer}
Ruibin Xiong, Yunchang Yang, Di He, Kai Zheng, Shuxin Zheng, Chen Xing,
  Huishuai Zhang, Yanyan Lan, Liwei Wang, and Tieyan Liu.
\newblock {On Layer Normalization in the Transformer Architecture}.
\newblock In {\em International Conference on Machine Learning (ICML)}, pages
  10524--10533. PMLR, 2020.

\bibitem{xu2017scene}
Danfei Xu, Yuke Zhu, Christopher~B Choy, and Li Fei-Fei.
\newblock {Scene Graph Generation by Iterative Message Passing}.
\newblock In {\em Proceedings of the IEEE Conference on Computer Vision and
  Pattern Recognition (CVPR)}, pages 5410--5419, 2017.

\bibitem{yang2018graph}
Jianwei Yang, Jiasen Lu, Stefan Lee, Dhruv Batra, and Devi Parikh.
\newblock {Graph R-CNN for Scene Graph Generation}.
\newblock In {\em Proceedings of the European Conference on Computer Vision
  (ECCV)}, pages 670--685, 2018.

\bibitem{yatskar2017commonly}
Mark Yatskar, Vicente Ordonez, Luke Zettlemoyer, and Ali Farhadi.
\newblock {Commonly Uncommon: Semantic Sparsity in Situation Recognition}.
\newblock In {\em Proceedings of the IEEE Conference on Computer Vision and
  Pattern Recognition (CVPR)}, pages 7196--7205, 2017.

\bibitem{yatskar2016situation}
Mark Yatskar, Luke Zettlemoyer, and Ali Farhadi.
\newblock {Situation Recognition: Visual Semantic Role Labeling for Image
  Understanding}.
\newblock In {\em Proceedings of the IEEE Conference on Computer Vision and
  Pattern Recognition (CVPR)}, pages 5534--5542, 2016.

\bibitem{you2016image}
Quanzeng You, Hailin Jin, Zhaowen Wang, Chen Fang, and Jiebo Luo.
\newblock {Image Captioning With Semantic Attention}.
\newblock In {\em Proceedings of the IEEE Conference on Computer Vision and
  Pattern Recognition (CVPR)}, 2016.

\bibitem{zhang2021spatially}
Frederic~Z. Zhang, Dylan Campbell, and Stephen Gould.
\newblock {Spatially Conditioned Graphs for Detecting Human-Object
  Interactions}.
\newblock In {\em Proceedings of the IEEE/CVF International Conference on
  Computer Vision (ICCV)}, pages 13319--13327, 2021.

\end{thebibliography}
